\documentclass[journal]{IEEEtran}

\usepackage[T1]{fontenc}    
\usepackage{hyperref}       
\usepackage{url}            
\usepackage{booktabs}       
\usepackage{amsfonts}       
\usepackage{nicefrac}       
\usepackage{microtype}      
\usepackage{lipsum}		
\usepackage{graphicx}
\usepackage{bbding}
\usepackage{amssymb}
\usepackage{doi}

\usepackage{array}
\usepackage{tabularx}
\usepackage{multirow}
\usepackage{cite}

\ifCLASSINFOpdf

\else

\fi

\begin{document}

%
\title{Exploring the Interplay Between Video Generation and World Models in Autonomous Driving: A Survey}
%
%
%

\author{Ao~Fu, Yi~Zhou,~\IEEEmembership{Senior Member,~IEEE,} Tao~Zhou,~\IEEEmembership{Senior Member,~IEEE,} Yi~Yang, Bojun~Gao, Qun~Li, Guobin~Wu, and Ling Shao~\IEEEmembership{Fellow,~IEEE}
\thanks{A. Fu and Y. Zhou are with the School of Computer Science and Engineering, Southeast University, Nanjing, China, and Key Laboratory of New Generation Artificial Intelligence Technology and Its Interdisciplinary Applications (Southeast University), Ministry of Education, China. Corresponding author: Yi Zhou (yizhou.szcn@gmail.com)}
\thanks{T. Zhou is with PCA Lab, and the School of Computer Science and Engineering, Nanjing University of Science and Technology, Nanjing, China.}
\thanks{Y. Yang, B. Gao, Q. Li and G. Wu are with DiDi Chuxing, Beijing, China}
\thanks{L. Shao is with the UCAS-Terminus AI Lab, University of ChineseAcademy of Sciences, Beijing, China}}

\maketitle
\begin{abstract}
World models and video generation are pivotal technologies in the domain of autonomous driving, each playing a critical role in enhancing the robustness and reliability of autonomous systems. World models, which simulate the dynamics of real-world environments, and video generation models, which produce realistic video sequences, are increasingly being integrated to improve situational awareness and decision-making capabilities in autonomous vehicles. This paper investigates the relationship between these two technologies, focusing on how their structural parallels, particularly in diffusion-based models, contribute to more accurate and coherent simulations of driving scenarios. We examine leading works such as JEPA, Genie, and Sora, which exemplify different approaches to world model design, thereby highlighting the lack of a universally accepted definition of world models. These diverse interpretations underscore the field’s evolving understanding of how world models can be optimized for various autonomous driving tasks. Furthermore, this paper discusses the key evaluation metrics employed in this domain, such as Chamfer distance for 3D scene reconstruction and Fréchet Inception Distance (FID) for assessing the quality of generated video content. By analyzing the interplay between video generation and world models, this survey identifies critical challenges and future research directions, emphasizing the potential of these technologies to jointly advance the performance of autonomous driving systems. The findings presented in this paper aim to provide a comprehensive understanding of how the integration of video generation and world models can drive innovation in the development of safer and more reliable autonomous vehicles.

\end{abstract}


%
\IEEEpeerreviewmaketitle

\section{Introduction}
%
%
%
%
\IEEEPARstart{W}{orld} models, which emerged from control theory and have progressively merged with reinforcement learning, facilitate the interaction between agents and their external environment, as well as simulate the dynamics of the world. These models enable agents not only to perceive and comprehend their surroundings but also to anticipate the unfolding dynamics of the world. In recent years, the integration of video generation techniques with world models has gained significant attention, particularly in the domain of autonomous driving \cite{drivedreamer1,drivedreamer2,drivingdiffusion,adriver,gaia1,mile,muvo,copilot4d}. 

Video generation tasks involve the automated creation of video content using machine learning models, particularly focusing on how these models can synthesize realistic and coherent video sequences. This task encompasses a variety of applications, from generating short video clips that match a given text description to extending existing videos or creating entirely new scenes based on learned data distributions.

At the core of video generation are specialized deep learning models that can understand and simulate the complex, dynamic nature of videos. These models must capture not only the appearance and style of objects within the video frames but also the temporal relationships and continuity that define realistic motion. Common approaches in video generation are exemplified by several techniques. Generative Adversarial Networks (GANs) \cite{gan} learn to produce video content that is indistinguishable from real videos through adversarial training. Another technique is Variational Autoencoders (VAEs) \cite{vae}, which model the probability distribution of input data to generate new instances. Additionally, diffusion models \cite{pvdm, LVDM, videocomposer, ho2022video, ho2022imagen, harvey2022flexible, yu2023video} transform noise into structured video sequences via a gradual denoising process. These approaches are demonstrated in works such as MoCoGAN \cite{tulyakov2017mocogan}, DIGAN \cite{yu2022DIGAN}, etc. 

The ultimate goal of video generation technologies is to produce videos that are not only visually pleasing but also maintain logical and temporal consistency across frames, mimicking the flow and evolution of scenes as seen in natural videos. As these technologies advance, they find applications in various fields such as entertainment, video games, virtual reality, and automated video editing and enhancement.

In the context of autonomous driving, video generation combined with world models is instrumental in enhancing the vehicle's ability to navigate complex environments. By generating realistic video sequences that simulate various driving scenarios, these models provide a robust framework for training and testing autonomous driving systems. They help in improving the vehicle's situational awareness and decision-making capabilities by predicting future states of the environment based on current observations.

The interplay between video generation and world models, particularly with a focus on diffusion models \cite{drivedreamer1,drivedreamer2,copilot4d}, presents a significant advancement in autonomous driving technologies. Diffusion models, known for their straightforward training regimen and high-quality output, have become a cornerstone in generative methodologies. These models integrate Langevin dynamics and stochastic differential equations to generate data, making them suitable for complex video generation tasks. The structural core of diffusion models aligns closely with the conceptual framework of world models. Both paradigms typically employ a two-stage process: an autoencoder for feature extraction and a core generative model for data synthesis. In diffusion models, the autoencoder compresses the data into a latent space, and the diffusion process then generates the final output by progressively refining the latent representation. Similarly, world models use perception modules to capture environmental data and prediction modules to forecast future states.

\begin{figure*}
	\centering
	\includegraphics[width=0.9\textwidth]{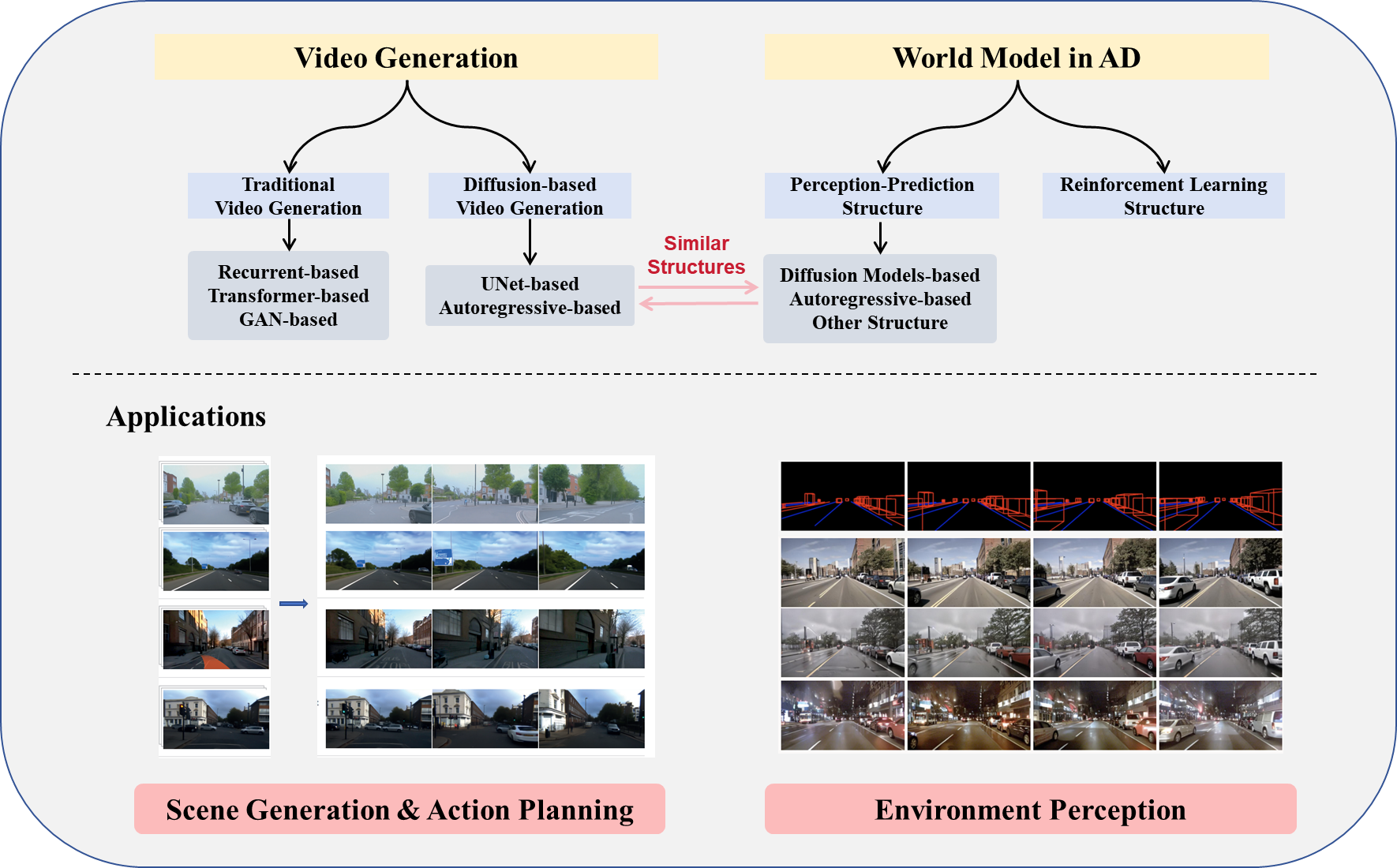}
	\caption{Video generation tasks can typically be divided into traditional video generation and diffusion-based video generation, the latter of which has seen widespread adoption in recent years. This paper categorizes world models in the autonomous driving domain into Perception-Prediction Structure and Reinforcement Learning Structure, with the former having a pipeline highly analogous to diffusion-based video generation tasks. This connection allows the paper to focus on the interplay between video generation and world models. Applications of world models in the autonomous driving field include Scene Generation, Action Planning, and Environment Perception. }
	\label{fig:fig1}
\end{figure*}

This synergy between diffusion models and world models enhances the capability of autonomous driving systems to simulate and predict real-world scenarios with high fidelity. The flexibility and control offered by diffusion models allow for the generation of diverse and realistic driving environments, crucial for the training and validation of autonomous vehicles. By leveraging these advanced generative techniques, autonomous driving systems can achieve more accurate and reliable performance in dynamic and complex road conditions.

\textbf{Comparisons with Related Surveys.} Several prior surveys have addressed the intersection of world models and autonomous driving, often focusing on broader themes such as end-to-end driving systems, perception, prediction, and control \cite{survey1, survey2, survey3}. However, these works generally lack detailed analysis of the interaction between video generation and world models. For example, existing surveys \cite{survey2} cover scenes generation and multimodal data integration but do not consider the specific interplay between video generation and world models. In contrast, our survey provides a focused review of the latest developments in this area, emphasizing the mutual enhancement potential of these technologies in complex autonomous driving scenarios. This approach offers a more specialized perspective, aiming to deepen the understanding of how video generation and world models can collaboratively advance autonomous driving systems.

\textbf{Contributions.} This survey explores the recent advancements and challenges in integrating video generation and world models, focusing on their applications in autonomous driving, which are illustrated in Figure \ref{fig:fig1}. It aims to highlight the structural similarities and synergistic potential of these technologies, providing insights into future research directions and practical implementations in the field of autonomous vehicle technology. Specifically, compared to other works, this review has the following four major contributions: 

\begin{itemize}

\item This survey analyzes the definitions of world models across various fields, emphasizing that the concept of a world model is not entirely fixed. It presents our understanding of world models and examines the structural uniformity of world models in the autonomous driving domain.
    
\item This survey highlights the structural similarities between video generation models and world models, explaining how these similarities enhance the performance and capabilities of autonomous driving systems.

\item It identifies key challenges and opportunities in integrating video generation and world models, providing insights into how these technologies can be further developed and applied in real-world scenarios.

\end{itemize}
\section{Video Generation}
\label{section2}
Video generation involves predicting future video frames by leveraging historical data through deep neural networks. The objective is to seamlessly extend a video sequence by accurately forecasting subsequent frames that align with the established visual and temporal dynamics. This task often incorporates conditional inputs, where future frames are not only generated based on prior frames but also influenced by supplementary conditions, such as text in text-to-video conversions. The process can be formally described as follows:
\[
\hat{x}_{t+1} = M(X, C)
\]
Here, for a given video sequence \(X = (x_0, x_1, ..., x_t)\), where \(x_i\) represents the i-th frame in the sequence, the next frame \(\hat{x}_{t+1}\) is predicted by the model \(M\), conditioned on \(X\) and additional variables \(C\).

\subsection{Traditional Video Generation}
In the initial stages, video generation technology struggled to produce natural, lengthy videos. Early models typically forecasted the next frame at a pixel level from patterns in training data or used probabilistic models to improve data distribution approximations. These attempts lacked a standardized structure, employing a variety of architectures such as Long Short-Term Memory (LSTM), Transformers, and Generative Adversarial Networks (GANs) to enhance generation quality through adversarial training. 
\textbf{Recurrent-based.} Recurrent-based networks handle the temporal dependencies in video sequences, which is crucial for understanding the dynamics over time. Their ability to remember long-term dependencies makes recurrent-based models suitable for scenarios with continuous actions or gradually evolving scenes, such as plot development in movies or surveillance video analysis. \cite{ConvLSTM} merged traditional Convolutional Neural Networks (CNNs) with recurrent networks, leveraging CNNs for extracting local features of individual frames and recurrent networks for understanding temporal and spatial dynamics between frames. \cite{predrnn} developed the Spatiotemporal LSTM (ST-LSTM), enhancing the temporal and spatial consistency in synthesized video sequences. While effective for capturing temporal dependencies, recurrent-based models can be computationally intensive and may struggle with very long sequences.


\textbf{Transformer-based.} Unlike recurrent-based models, Transformers can process entire data sequences at once, increasing complexity and expressive power by stacking more layers. The VideoGPT, introduced by \cite{videogpt}, combines 3D convolutional networks with Transformers to form a GPT-like structure, utilizing attention mechanisms to focus on entire video sequences. To address high computational costs, \cite{cogvideo} developed CogVideo, leveraging a pre-trained text-to-image model and proposing a multi-frame-rate hierarchical training strategy to align text with video clips. Transformers are highly effective at modeling long-range dependencies but often require substantial computational resources, making them less practical for real-time applications.

\textbf{GAN-based.} Through adversarial training, GANs create novel and realistic video content. Models like \cite{pose} incorporated human posture maps as prior knowledge, using a Variational Autoencoder (VAE) for feature extraction and GANs for enhancing realism. \cite{yu2022DIGAN} introduced DIGAN, employing Implicit Neural Representations (INRs) to enhance spatiotemporal consistency. \cite{tulyakov2017mocogan} proposed MoCoGAN, decoupling videos into content and motion, enabling random generation through unsupervised training and separate latent spaces. While GANs generate high-quality outputs, they are challenging to train and may encounter issues like mode collapse, which can reduce their flexibility.

Traditional video generation models, while pioneering, face significant limitations affecting their practical application. They often lack generalizability and adaptability across different video types, and the quality of generated videos frequently falls short, particularly in maintaining natural-looking, coherent long sequences. These challenges stem from limitations in model architecture and the computational burden of training complex networks, highlighting the need for innovative approaches. The next section explores advanced video generation techniques based on diffusion models, aiming to enhance quality, coherence, and length of generated videos, setting a new standard for this field.

\subsection{Diffusion-based Video Generation}
\subsubsection{Diffusion Models for Image Generation}
The diffusion model \cite{DDPM,DDIM, saharia2022palette, xia2023diffir, zhang2023adding, aboulaich2008new, gu2022vector}, a probabilistic generative model that integrates Langevin dynamics and stochastic differential equations, has emerged as a cornerstone in generative methodologies. Due to its straightforward training, high-quality output, and control capabilities, it has been widely adopted. Contemporary SOTA video generation models rely heavily on the diffusion model framework. The denoising diffusion probabilistic model (DDPM) \cite{DDPM} represents the most intuitive and broadly accepted formulation. Training a diffusion model involves two stages: the forward process and the reverse process. In the forward process, noise is incrementally added to the image until it becomes nearly indistinguishable from Gaussian noise. In the reverse process, the model systematically reconstructs the original image from this noise.

Formally, \(x_0\) is defined as the distribution of the training data, denoted by \(p_{\theta}(x_0) := \int p_{\theta}(x_{0:T}) dx_{1:T}\). In the forward process, \(x_0\) is incrementally corrupted into \(x_t\), where \(x_1, ..., x_T\) are the intermediate latent states represented by the distribution:
\[
q(x_{1:T}|x_0) := \prod_{t=1}^{T} q(x_t|x_{t-1})
\]
\[
q(x_t|x_{t-1}) := \mathcal{N}(x_t; \sqrt{1-\beta_t}x_{t-1}, \beta_t I)
\]
Here, \(\beta_t\) is a hyperparameter guiding the noise addition process.

The reverse process is a Markov chain with learned Gaussian transitions, starting from \( p(x_T) = \mathcal{N}(x_T; 0, I) \):
\[
p_{\theta}(x_{0:T}) := p(x_T) \prod_{t=1}^{T} p_{\theta}(x_{t-1}|x_t)
\]
\[
p_{\theta}(x_{t-1}|x_t) := \mathcal{N}(x_{t-1}; \mu_{\theta}(x_t, t), \Sigma_{\theta}(x_t, t))
\]
This process involves iterative calculation based on the corresponding distributions, gradually reconstructing the image from Gaussian noise. 

To address computational intensity, \cite{LDM} established the Latent Diffusion Model (LDM), using an Auto-Encoder structure to downsample data. By incorporating KL divergence or VQ regularization \cite{vqvae}, data is compressed into a manageable latent space, significantly accelerating the generation process without compromising quality. Additionally, LDM leverages pre-trained encoders from various fields as condition extractors, allowing the diffusion model to be guided by a wide range of prompts, enhancing flexibility and control. \cite{DiT} developed the Diffusion Transformers (DiT), replacing the commonly used UNet module with a Transformer for better scalability and processing of large datasets. Images are tokenized and recursively denoised within the Transformer, mimicking the traditional diffusion model's iterative noise reduction.

Diffusion models have incorporated various components, significantly enhancing their generative capabilities, and have established a solid foundation for tasks in video generation.

\subsubsection{Diffusion Models for Video Generation}
Due to the ease of training diffusion models and their general structure, research on video generation often follows a fixed pipeline architecture, distinct from previous works. This pipeline typically comprises the autoencoder and the core diffusion model. The training process is divided into two stages: \textbf{(i)} The autoencoder learns feature representations of the data, compressing it into a latent space. \textbf{(ii)} The diffusion model is then trained within this latent space, focusing on generating content by leveraging the compressed feature representations.
\begin{figure}
	
	\includegraphics[height=0.45\textwidth, width=0.45\textwidth]{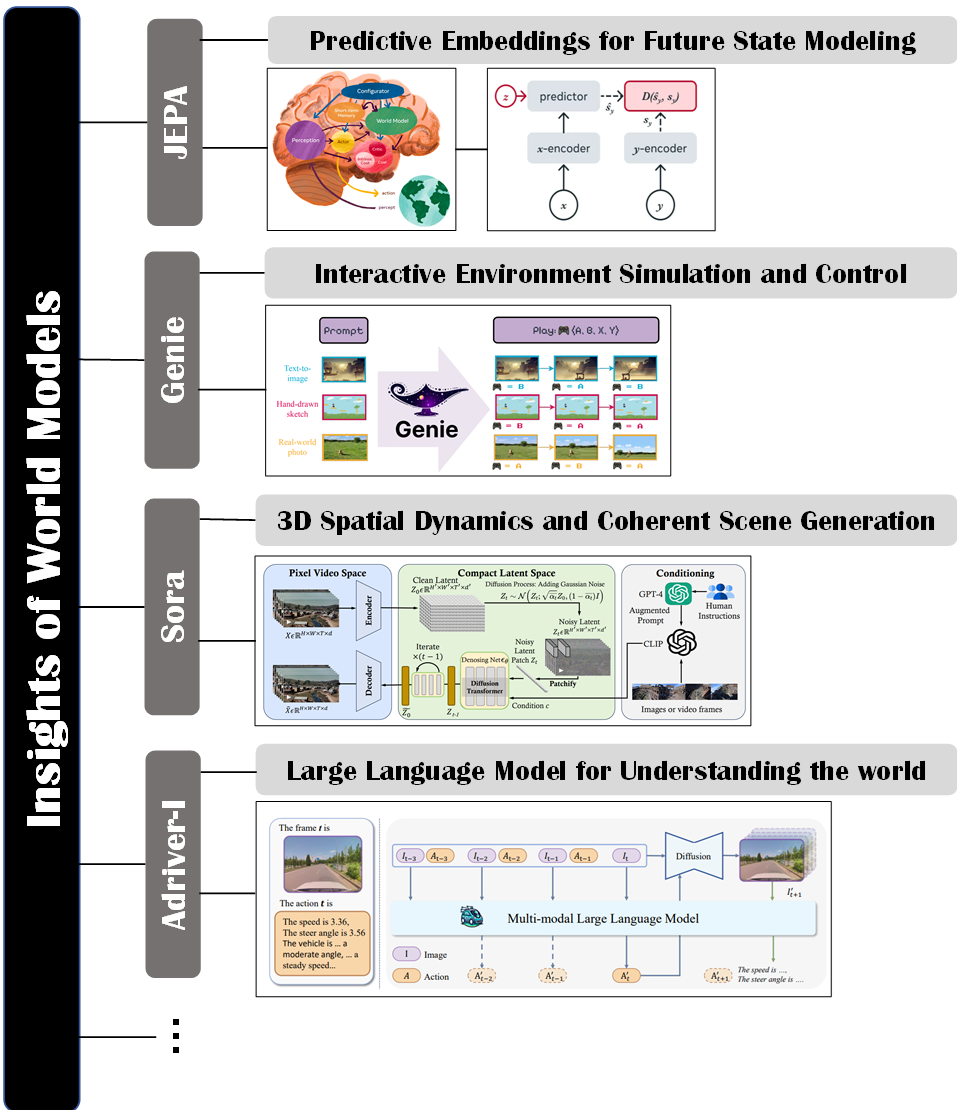}
	\caption{The interpretations and pipeline designs of world models presented in the works JEPA, Genie, Sora and ADriver-I indicate that the concept of world models lacks a universally accepted definition. These three works not only highlight the diversity of thought within the field but also represent distinct approaches to understanding and implementing world models, each reflecting different conceptualizations and priorities in model design.
}
	\label{fig:fig4}
\end{figure}
\textbf{UNet-based:} Diffusion models typically employ UNet due to its capability for multi-scale feature extraction and detail capture facilitated by skip connections. \cite{pvdm} introduced the Projected Latent Video Diffusion Model (PVDM), projecting video data into three dimensions to obtain 2D feature maps, which are processed by the diffusion model and reconstructed back into video space. \cite{LVDM} proposed the Latent Video Diffusion Models (LVDM), utilizing a conditional 3DUnet and a hierarchical diffusion method for tasks like text-to-video synthesis and long-duration video generation. \cite{videocomposer} engineered VideoComposer, integrating it with a Spatio-Temporal Condition Encoder (STC-encoder) for multi-conditional video generation.

\textbf{Autoregressive-based:} The scalability of models and their ability to process large datasets are crucial in video generation tasks. \cite{magvit} introduced the MAsked Generative VIdeo Transformer (MAGVIT), utilizing a 3D tokenizer to serialize video inputs and combining techniques from MaskGIT \cite{maskgit} and diffusion models. Copilot4d \cite{copilot4d} is a scenario generation model for autonomous driving, predicting future 3D representations of a scene. It integrates a tokenizer with a Transformer architecture, facilitating the prediction of environmental dynamics and modeling the agent's comprehension of the surrounding world.

\section{World Models in AD}
\subsection{Basics of the World Model}

World models, which emerged from control theory in the 1970s, have progressively merged with the discipline of reinforcement learning. The core objective of these models is to forge a linkage between an agent and its operational environment utilizing what are termed 'world models.' This conceptual framework mandates that agents must have the capacity to not only perceive and comprehend their surroundings but also to anticipate the unfolding dynamics of the world. This integration highlights the expanding scope of reinforcement learning applications, extending its utility in complex decision-making scenarios.

\begin{figure*}
	\centering
	\includegraphics[width=0.9\textwidth]{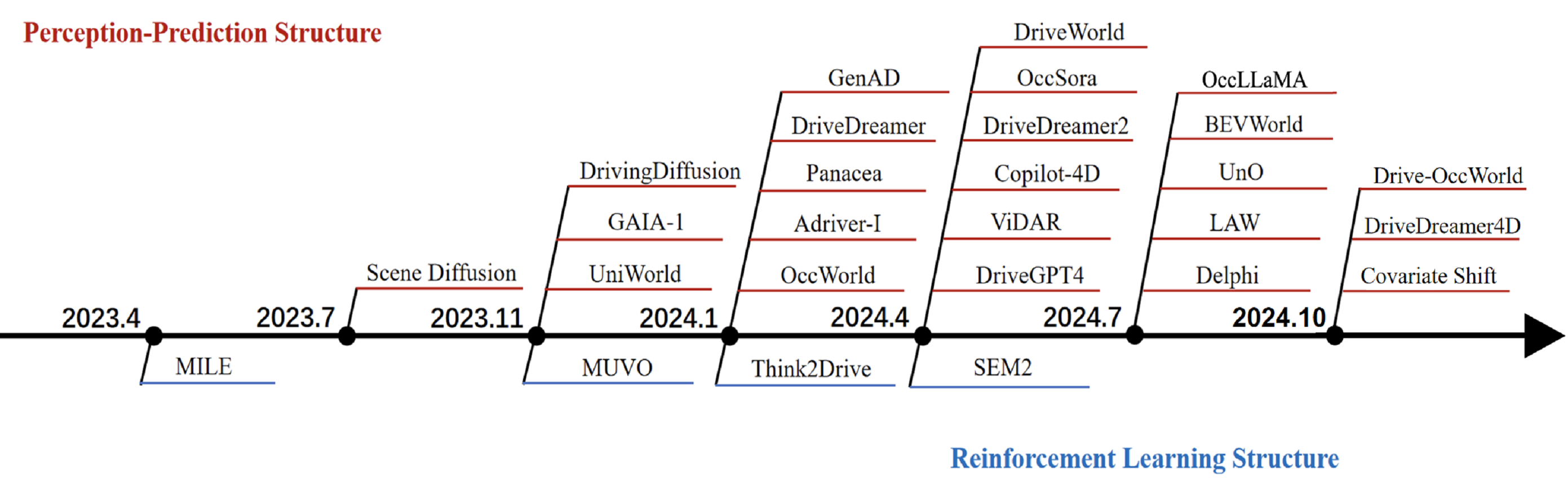}
	\caption{Timeline of World Models in Autonomous Driving. Based on the key architectures, we categorize the models from various studies into Perception-Prediction structures and Reinforcement Learning structures. This paper primarily focuses on the former, as its interplay with video generation models is more prominent.}
	\label{fig:fig2}
\end{figure*}

Reinforcement learning algorithms inherently demand extensive iterative computations to reach optimal solutions, which makes the use of such algorithms for world simulations particularly challenging. With the evolution of deep neural networks, \cite{worldmodel} have successfully merged reinforcement learning with these networks to effectively simulate scenarios within video games, effectively breaking down the conventional frameworks of reinforcement learning. While interactions between agents and environments, along with agent training optimization, still leverage traditional reinforcement learning methods, the evolution of the "world" is now modeled through a neural network. This network receives the current perception of the agent and outputs corresponding actions or reactions based on the current state. Specifically, they employ a VAE to learn the data distribution, acting as the agent's sensory system for gathering and representing environmental information. Moreover, they combine a Mixture Density Network (MDN) \cite{MDN} with a recurrent neural network (RNN) \cite{RNN} to create an MDN-RNN module. This module serves as the agent's processor, capturing and interpreting temporal data, thus equipping the model with predictive capabilities concerning future developments.

Inspired by the concept of world models, \cite{PlaNet} proposed PlaNet, a framework based on reinforcement learning that combines RNN to design the Recurrent State-Space Model (RSSM). This model learns environmental dynamics from images and selects actions through rapid online planning in the latent space. Building on this foundation, several similar works have emerged, such as the Dreamer series \cite{Dreamer1,Dreamer2,Dreamer3}. These works aim to solve long-term tasks from images by leveraging latent imagination. They propagate analytic gradients of learning state values through imagined trajectories in the compact state space learned by the world model, effectively learning behaviors.

Yann LeCun, a seminal figure in deep learning, asserts that both humans and animals inherently possess the ability to understand the workings of the world through what is commonly referred to as "common sense." This profound level of understanding is critically needed by modern artificial intelligence models to truly grasp the dynamics of the world. To address this gap, LeCun introduced a novel AI architecture specifically tailored for autonomous learning and comprehensive understanding of the world \cite{lecun2022path}. The core of this architecture is the "world model" module, designed to predict future states of the world, supported by auxiliary modules such as the Configurator and Perception. Aligned with this innovative framework, LeCun and his team have crafted the Image-based Joint-Embedding Predictive Architecture (I-JEPA) \cite{ijepa}, which distinguishes itself from traditional generative models by applying its loss function within the embedding space instead of the raw data space. This strategic focus on predicting the embedding representations of input data prioritizes learning abstract representations rather than mere direct reconstruction. Building on the JEPA framework, the team further developed the Video Joint Embedding Predictive Architecture (V-JEPA) \cite{vjepa} and Image World Models (IWM) \cite{IWM}, thus enhancing the conceptualization and application of world models.

OpenAI and Google have each unveiled their respective innovations, Sora \cite{sora} and Generative Interactive Environments (Genie) \cite{genie}, showcasing their unique interpretations of world models. Sora, developed by OpenAI, is a large-scale video generation model based on the DiT architecture. It has been trained on extensive video datasets, demonstrating impressive capabilities, particularly in generating long videos, leading its creators to view it as a model capable of simulating the real world. This model stands out for its exceptional ability to generate videos, including extended footage, showcasing its potential to mimic complex real-world dynamics. However, it still falls short in accurately simulating certain physical laws such as fluid dynamics and causal relationships. On the other hand, Google's Genie represents the first generative interactive model trained using unsupervised video data. It focuses on decoupling scenes and actions to simulate game environment construction. Upon receiving action commands, Genie can generate outcomes reflecting the protagonist's responses in the game world, positioning it as a world model for gaming spaces. Closer to the original concept of world models, Genie simulates an agent that interacts with its environment. Although it is good at promoting interactive scenes, its overall fidelity in terms of visual representation and scene integration is not satisfactory enough.

\subsection{World Model for Autonomous Driving}

From reinforcement learning-centric world models to JEPA, which focuses on data abstraction, to the data-driven DiT-based Sora known for its advanced generative capabilities, and Genie, which simulates game scenarios—these diverse models are all defined as world models. In contrast to this diverse array of world models, those in the domain of autonomous driving often exhibit a unified structure: a perception module and a prediction module. The perception module serves as an intermediary between the model and its external environment, compressing incoming data into a specific representational format that eases the burden on the subsequent prediction process. The prediction module uses this refined data to forecast future states, which could encompass environmental scenes, decision-making processes.

\begin{figure*}
	\centering
	\includegraphics[width=\textwidth]{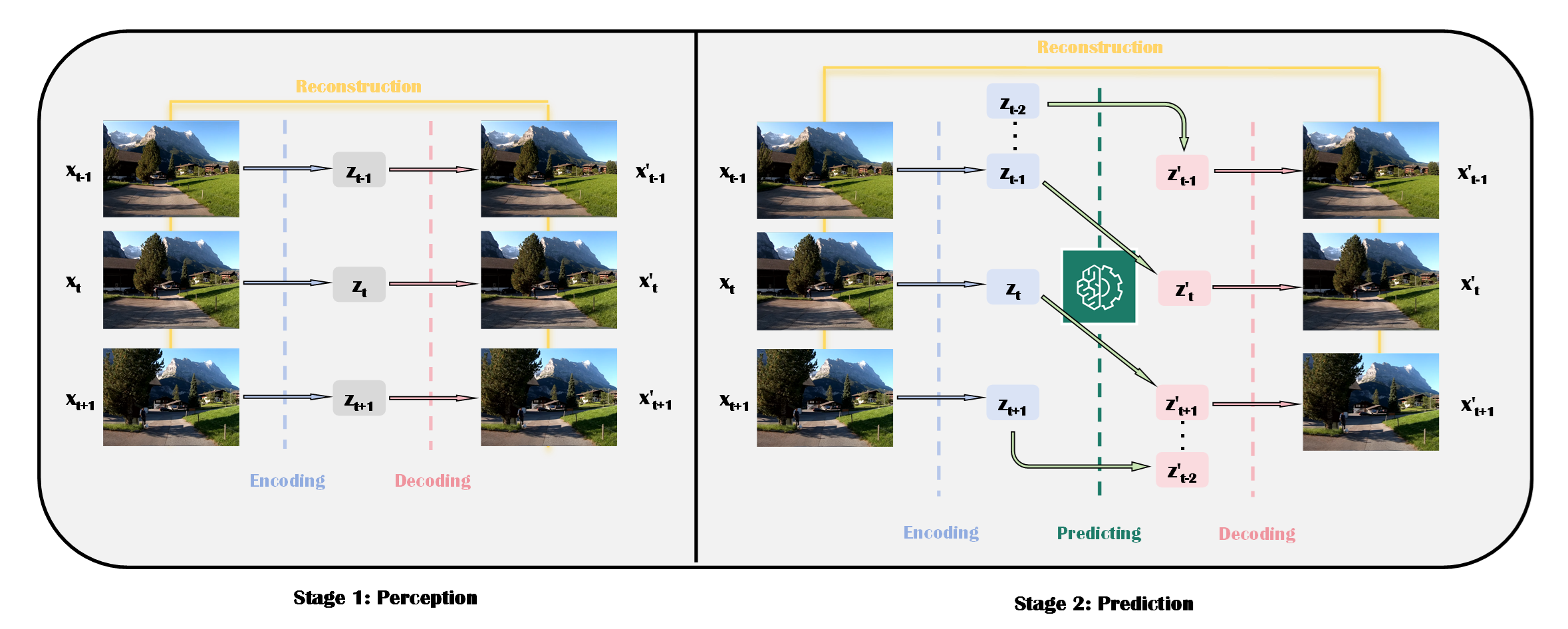}
	\caption{The unified architecture of video generation and world models in autonomous driving. {\includegraphics[width=0.03\textwidth]{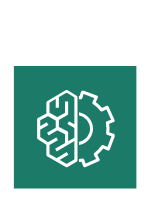}} is generative models or world models with temporal prediction capabilities. Both methodologies encompass two primary components: the perception and simulation module and the core prediction module. The perception and simulation module is primarily responsible for learning data distributions, extracting real-world data.  Conversely, the prediction module focuses on learning the dynamic patterns of data within the compressed space, thereby simulating data changes. }
	\label{fig:fig3}
\end{figure*}

Similarly, the architectural foundation of video generation frameworks based on diffusion models also splits into two main components: an autoencoder that captures and decodes data patterns and a core diffusion model that predicts data distributions, which are illustrated in Figure \ref{fig:fig3}. This structural design ensures that the models can effectively process and generate complex data. The structural similarities between autonomous driving world models and diffusion-based video generation frameworks highlight the effectiveness of this architectural approach. Leveraging these advanced models enhances situational awareness and decision-making capabilities in autonomous driving systems, paving the way for more reliable and efficient autonomous vehicles. Therefore, this section will categorize common world models in the field of autonomous driving according to the aforementioned structure, TABLE \ref{tab:overview} and Figure \ref{fig:fig2} summaries the methods with world models in Autonomous Driving.

\subsubsection{Perception-Prediction Structure}

As discussed earlier, most world models in the field of autonomous driving are based on fixed structures. Whether it is a multimodal encoder or individual modality tokenizers, they serve as the link between the environment and the model, acting as the model's perceivers to gather information and extract features. Both diffusion models, which excel at handling visual information, and Transformer structures, which are adept at processing serialized and linguistic information, are used to fit and predict real-world data distributions. Therefore, this paper categorizes such structures as Perception-Prediction Structures.

\textbf{Diffusion-Based:}
Recent advancements in autonomous driving are markedly influenced by the integration of diffusion models, enhancing predictive capabilities and realism in simulation environments. This diffusion model-centric architecture places a significant emphasis on visual perceptual information. These models compress all information into a latent space using various methods, leveraging the superior generative capabilities of diffusion models to simulate the world. From a visual perspective, they predict the evolution of the environment, treating the model as a simulator for the real world. DriveDreamer series (DriveDreamer and DrvieDreamer-2) innovate by deriving world models from extensive real-world driving data, allowing for predictive simulations and enhanced policy-making in autonomous driving applications. \cite{drivedreamer1} introduce multimodal information, such as depth maps, RGB images, HD maps, and text, through various encoders. The core diffusion model then learns the gradients of these inputs with respect to random Gaussian noise. Finally, different decoders output the predicted information for the next time step, providing a comprehensive simulation of future states. From structured traffic constraints to user-defined scenario simulations, DriveDreamer-2 \cite{drivedreamer2} demonstrates substantial advancements in generating controllable and realistic driving videos, supporting both current and future autonomous driving tasks. Especially, by structuring traffic simulations around user inputs via text prompts, DriveDreamer-2 dynamically generates foreground (agent trajectories) and background (HDMaps) conditions, feeding them into a unified multi-view video model (UniMVM) for cohesive video synthesis. This approach not only improves the fidelity and consistency of the generated videos but also demonstrates substantial advancements in training perception methods for autonomous driving, confirmed by improved detection and tracking metrics in experimental evaluations.

Driving Diffusion \cite{drivingdiffusion} with a similar structure presents a sophisticated framework for generating multi-view videos from 3D layouts in complex urban scenes. They separately train structurally similar multi-view and temporal models. During the inference phase, the two models are concatenated. The multi-view model receives multimodal information and generates multi-view keyframes to serve as reference frames. The temporal model then interpolates frames based on these multi-view reference frames to generate the video. It utilizes temporal attention blocks and consistency loss to ensure continuity between frames. This framework, known as Driving Diffusion, addresses the challenges of maintaining cross-view and cross-frame consistency and improving the quality of generated instances such as vehicles and pedestrians. This novel approach allows for the creation of large-scale, realistic multi-camera driving videos without additional costs, significantly benefiting downstream driving tasks. The Scene Diffusion system \cite{scenediffusion} extends this concept by using diffusion processes to generate discrete bounding boxes that mimic a self-driving car's perception system, adapting effectively to varied geographic traffic patterns. This technique not only simplifies the generation process but also enhances the model’s ability to produce diverse traffic scenarios that reflect real-world conditions. The paper demonstrates the system’s robust performance across multiple datasets, showing significant promise for future applications in autonomous driving and advanced traffic management systems. Panacea \cite{panacea} introduces a novel multi-temporal autonomous driving scenario generation model based on diffusion models. By integrating state-of-the-art generative models such as ControlNet\cite{contorlNet}, Panacea aims to enhance both consistency and controllability. High consistency ensures that deformations between consecutive generated frames are significantly reduced, thereby improving inter-frame correlation. High controllability allows users to generate highly diverse results tailored to their specific needs. Additionally, this work plans to filter and integrate the generated data, releasing a dataset titled "Gen-nuScenes," to provide a platform for deeper research in the autonomous driving domain. GenAD \cite{GenAD1} is a comprehensive video prediction model tailored for autonomous driving applications. The model training process is bifurcated into two distinct stages: initially, a pre-trained latent diffusion model named SDXL is fine-tuned using driving images; subsequently, temporal reasoning blocks are integrated to bolster the model's capability to forecast future frames based on past observations, thereby addressing challenges such as causal reasoning and significant view shifts. The methodology encompasses the creation of the OpenDV-2K dataset, comprising over 2000 hours of driving videos sourced from YouTube and various public datasets, enriched with textual descriptions.

In the task of generating autonomous driving scene videos using world models, the precision and long-term stability of video generation are paramount. Delphi \cite{Delphi} and VISTA \cite{Vista} both make substantial contributions to the quality of world model generation in autonomous driving, focusing on fine-grained features and the length and stability of generated videos. Delphi introduces a diffusion-based method that tackles spatial and temporal inconsistencies in long video generation. It incorporates a Noise Reinitialization Module (NRM) and a Feature-aligned Temporal Consistency Module (FTCM). These innovations enable Delphi to generate up to 40 frames of consistent multi-view videos, significantly surpassing previous methods. Moreover, Delphi employs a failure-case driven framework to enhance model performance by generating training data akin to identified failure cases, thereby improving robustness and reducing collision rates in end-to-end autonomous driving models. Conversely, VISTA emphasizes the integration of prior knowledge to improve video generation quality for autonomous driving. It leverages multi-modal data inputs and advanced techniques to ensure long-term temporal consistency in the generated videos. VISTA’s approach incorporates structured priors into the generative process, maintaining high fidelity and stability in output sequences. This method enables VISTA to produce high-quality, diverse video sequences essential for training reliable autonomous systems.

To shows significant improvements in point cloud forecasting, reducing prior state-of-the-art Chamfer distances substantially across multiple datasets. Copilot4D  \cite{copilot4d} is an innovative approach where sensor observations are tokenized using a VQVAE model, followed by future prediction through discrete diffusion. Different from previous methods,  Copolit4D is mainly applied to the generation of 3D point cloud scenes. As mentioned above, this generation task is similar to video generation, and both add time dimensions to the original 2D pictures or 3D point clouds. They explores advancing unsupervised learning in autonomous driving by leveraging discrete diffusion processes. Analogously, OccSora\cite{OccSora} introduces a diffusion-based 4D occupancy generation model for dynamically simulating 3D worlds in autonomous driving. The key contributions include an innovative framework that efficiently simulates long-term scene evolution, surpassing traditional autoregressive methods. The model employs a 4D scene tokenizer to compress and reconstruct 4D occupancy data into compact spatio-temporal representations. It then utilizes a diffusion transformer to learn from these representations and generate 4D occupancy conditioned on a trajectory prompt. The diffusion process involves both forward noise introduction and reverse propagation to refine scene representations, leveraging token embeddings, trajectory conditioning, and noise handling. By using multidimensional diffusion techniques, OccSora accurately propagates 4D information, ensuring temporal consistency and realistic scene evolution. Extensive experiments on the nuScenes dataset demonstrate its ability to generate realistic 16-second 3D layout videos, showcasing the model's robustness and adaptability to various trajectory inputs, making it a significant advancement in autonomous driving simulation.

Beyond conventional RGB pixel-space video data and 3D representations such as occupancy grids and point clouds, training with Bird's Eye View (BEV) data is gaining traction. BEVWorld \cite{BEVWorld} creates a unified BEV latent space that integrates multimodal sensor data, marking significant progress in this field. The main contributions of BEVWorld include a multimodal tokenizer that encodes visual and LiDAR data into a compact BEV representation and a latent BEV sequence diffusion model that predicts future scenes based on action tokens. The tokenizer uses an autoencoder structure to compress multimodal data, achieving self-supervised reconstruction of high-resolution images and point clouds through ray-casting rendering. The diffusion model employs a spatiotemporal transformer to convert noisy BEV tokens into clean future predictions, thereby avoiding the cumulative errors associated with autoregressive methods. Extensive experiments on the nuScenes and Carla datasets demonstrate BEVWorld's superior performance in generating accurate future scenes and enhancing downstream tasks such as perception and motion prediction, signifying a major advancement in autonomous driving technology.

\textbf{Autoregressive-Based:}
With the significant success of data-driven approaches, the potential of Transformers has been further realized. Unlike models centered around diffusion models to fit the world, many world model works have focused on the powerful sequential processing and autoregressive prediction capabilities of Transformers, as well as the mature and well-established pre-trained Large Language Models (LLMs) and the highly capable Multimodal Large Language Models (MLLMs). As one of the earliest proposed world models, GAIA-1\cite{gaia1} utilizes Transformers as the core component for world modeling. It processes multimodal inputs such as video, text, and actions, mapping them into a discrete token space. By leveraging the superior sequence modeling and prediction capabilities of Transformers, GAIA-1 ultimately decodes predictions to generate realistic driving scenarios. During the encoding phase, inputs like video, text, and actions are encoded and concatenated to serve as the input at a given moment, thereby predicting the corresponding information for the next moment. GAIA-1 successfully treats world modeling as an unsupervised sequence modeling problem, achieving significant results. ViDAR \cite{vidar} is a cutting-edge visual autonomous driving pre-training model that leverages historical visual inputs to predict future point clouds, significantly enhancing perception, prediction, and planning tasks. The model addresses the limitations of existing methods by incorporating a novel architecture consisting of a history encoder, a latent rendering operator, and a Transformer-based future decoder. By synergistically capturing semantics, 3D geometry, and temporal dynamics, ViDAR achieves superior performance across various downstream applications. Experimental results demonstrate that ViDAR markedly outperforms traditional approaches, offering higher accuracy and efficiency in autonomous driving scenarios. 

Pre-trained LLMs \cite{wu2023next, ma2023llm, ni2023lever, vaithilingam2022expectation, ding2023static, gu2023llm}, such as LLaMA \cite{llama, llama2}, Vicuna \cite{vicuna, vicuna_} and GPT \cite{gpt,gpt2,gpt3, floridi2020gpt}, possess a broad knowledge base and exhibit strong language understanding capabilities. Building upon this foundation, MLLMs, such as LLaVA \cite{llava}, BLIP2 \cite{blip2} and GPT4 \cite{gpt4}, can handle and understand multiple data types, generating cross-modal results. Consequently, in the field of autonomous driving, there are numerous world models based on LLMs and MLLMs. These models typically tokenize the training data and process it using pre-trained models. This approach treats all data as serialized linguistic information, extracting relevant semantic features to simulate and predict the world. For example, ADrive-I \cite{adriver} is a multi-modal large language models with diffusion techniques. It uses interleaved vision-action pairs to unify visual features and control signals, facilitating direct output of control signals and prediction of future scenes. This integrated approach enhances simulation realism and dynamic interaction by predicting continuous actions and generating corresponding visual frames. Tested on extensive datasets, ADriver-I demonstrates superior performance in prediction accuracy and scene generation, offering new directions for autonomous driving technologies. \cite{drivegpt4} presents DriveGPT4, an end-to-end large language model for autonomous driving. This work leverages a pre-trained MLLM to tokenize various multimodal data inputs into the model. The result is a model that predicts current vehicle control signals and answers user queries, creating an application-oriented large language model. Building on recent advancements in world models, notable progress has been made in integrating language with action through occupancy-based generative world models. A prominent example is OccLLLaMA \cite{wei2024occllama}, which leverages semantic occupancy as a general visual representation while unifying vision, language, and action modalities. By employing a VQVAE-like scene tokenizer, OccLLLaMA discretizes and reconstructs occupancy scenes, capturing sparse environments with high efficiency. This model enhances multi-modal predictions through the use of a unified vocabulary, further improving its capacity for 4D occupancy forecasting, motion planning, and question answering in the context of autonomous driving tasks.

The evolution of 3D scenes more accurately reflects the visual scenarios encountered in autonomous driving. To enhance the predictive capability, efficiency, and versatility of 3D scene prediction, \cite{occworld} proposes that models learn based on 3D occupancy, which can describe finer-grained 3D structures. They introduce a world model named OccWorld. This model first pre-trains the tokenizer using conventional reconstruction tasks, then employs a GPT-like LLM as the core generative model. Supplemented by a temporal causal self-attention module, OccWorld improves the consistency of predictive results and reduces the occurrence of unavoidable physical law anomalies. Previous works have largely focused on integrating predictive modeling with trajectory planning, but often lacked a comprehensive approach to handle complex driving environments effectively. Drive-OccWorld \cite{driveoccworld} integrates future occupancy prediction with end-to-end trajectory planning, presenting a novel vision-centric world model for autonomous driving. By incorporating controllable action-conditioned generation and continuous occupancy state forecasting, Drive-OccWorld significantly enhances decision-making in autonomous systems, offering increased safety, interpretability, and adaptability to complex driving environments. 

In the 4D space formed by adding a temporal dimension to 3D data, Uniworld \cite{uniworld} utilizes Transformers to fit the data. It employs unsupervised pre-training with LiDAR data to construct a 4D spatiotemporal model, which is then fine-tuned for various downstream tasks to enhance performance across different applications. This unified pre-training framework has demonstrated impressive results in critical tasks such as motion prediction, multi-camera 3D object detection, and surrounding semantic scene completion. Compared to monocular pre-training methods on the nuScenes dataset, Uniworld significantly improved IOU for motion prediction by approximately 1.5\%. In multi-camera 3D object detection, MAP and NDS increased by 2.0\% each, and MIOU for semantic scene completion saw a 3\% rise. 

Scene simulation and reconstruction in autonomous driving have consistently posed significant challenges. Existing methods, such as NeRF \cite{nerf} and 3D Gaussian Splatting (3DGS) \cite{3d}, primarily focus on forward-driving scenarios based on training data distribution, leading to limitations when dealing with complex vehicular maneuvers such as lane changes, acceleration, and deceleration. To address these shortcomings, DriveDreamer4D \cite{drivedreamer4d} introduces a generative mechanism grounded in world models, aiming to improve the 4D representation and synthesis of driving scenes. This approach provides more detailed visual generation and dynamic modeling, especially in intricate driving scenarios, advancing the state-of-the-art in autonomous vehicle scene simulation.

\newcolumntype{C}{>{\centering\arraybackslash}X}

\begin{table*}
\centering
\caption{Summary of Structure, Methods and Core Architectures}
\begin{tabularx}{\textwidth}{CCCC}

\hline
\textbf{Structure} & \textbf{Method} & \textbf{Core Architecture} & \textbf{Data Type}\\
\hline
\\
\multirow{20}{*}{Perception-Prediction Structure} 
& DriveDreamer\cite{drivedreamer1} & Diffusion Model & 2D RGB \\
& DriveDreamer-2\cite{drivedreamer2} & Diffusion Model & 2D RGB\\
& DrivingDiffusion\cite{drivingdiffusion} & Diffusion Model & 2D RGB\\
& SceneDiffusion\cite{scenediffusion} & Diffusion Model & 2D RGB\\
& Panacea\cite{panacea} & Diffusion Model & 2D RGB\\
& GenAD\cite{GenAD1} & Diffusion Model & 2D RGB\\
& Delphi\cite{Delphi} & Diffusion Model & 2D RGB \\
& Vista\cite{Vista} & Diffusion Model & 2D RGB \\
& DriveDreamer4D\cite{drivedreamer4d} & Diffusion Model&2D RGB\\
& CoPolit4D\cite{copilot4d} & BEV Pillar Pooling + DiT & 3D Point Cloud \\
& OccSora\cite{OccSora} & Diffusion Model & 3D Occupancy \\
& BEVWorld\cite{BEVWorld} & Diffusion Model & 2D RGB / 3D Point Cloud\\
& GAIA-1\cite{gaia1} & Transformer & 2D RGB\\
& ViDAR\cite{vidar} & Transformer & 3D Point Cloud\\
& Uniworld\cite{uniworld} & Transformer  & 2D RGB / 3D Occupancy\\
& LAW\cite{LAW} & Transformer & 2D RGB\\
& Covariate Shift\cite{mitigating} & Transformer + MLP &2D RGB\\
& ADriver-I\cite{adriver} & MLLM & 2D RGB \\
& DriveGPT4\cite{drivegpt4} & MLLM & 2D RGB \\
& OccWorld\cite{occworld} & GPT-like LLM & 3D Occupancy \\

& UnO\cite{UnO} & AutoEncoder & 3D Point Cloud\\
& Drive-OccWorld\cite{driveoccworld} & AutoEncoder &3D Occupancy\\
& DriveWorld\cite{DriveWorld} & MSSM &3D Occupancy\\

\\
\hline

\\
\multirow{4}{*}{Reinforcement Learning Structure} 
& Thin2Drive\cite{think2drive} & RSSM & Action Text\\
& SEM2\cite{sem2} & RSSM &2D RGB\\
& MUVO\cite{muvo} & GRU &2D RGB / 3D Occupancy\\
& MILE\cite{mile} & PGM & 2D RGB\\
\\
\hline
\end{tabularx}
\label{tab:overview}
\end{table*}
To address the prohibitive annotation cost of high-dimensional data, a novel self-supervised learning approach, termed the LAtent World (LAW) \cite{LAW} model, is introduced for end-to-end autonomous driving. This approach aims to enhance driving performance by predicting future latent features based on the predicted ego actions and the latent features of the current frame. These predicted features are supervised by the actually observed features, enabling joint optimization of latent feature learning and action prediction. The proposed view selection via latent substitution strategy dynamically selects informative views for feature extraction while substituting unprocessed views with predicted latent features from the latent world model. This strategy involves predicting rewards for various view selection strategies, selecting the strategy with the highest reward, and using the selected views to predict the vehicle's trajectory. This method significantly enhances the efficiency of the processing pipeline with minimal performance loss. The network architecture employs a transformer for self-attention in the view dimension to facilitate these predictions.

\textbf{Other Structure}
In addition to conventional approaches based on diffusion models and Transformer-based autoregressive sequence processing, some works introduce different model architectures from other perspectives. 

UnO \cite{UnO} presents a novel unsupervised method for predicting 4D occupancy fields from LiDAR data. This model employs a simple ResNet-based autoencoder structure to learn BEV feature maps and an implicit occupancy decoder for continuous occupancy prediction across space and time. By generating pseudo-labels through sampling positive and negative examples, UnO can accurately predict geometric structures, dynamics, and semantics without requiring expensive annotations. It significantly outperforms state-of-the-art methods in point cloud forecasting and BEV semantic occupancy prediction, demonstrating its effectiveness and transferability to various autonomous driving tasks. A novel 4D pre-training framework named DriveWorld \cite{DriveWorld} enhances scene understanding for autonomous driving by leveraging world models. This approach employs a unique structure, the Memory State-Space Model (MSSM), which integrates a Dynamic Memory Bank for temporal dynamics and a Static Scene Propagation module for spatial context. Adaptive feature extraction is facilitated by a Task Prompt mechanism tailored to various tasks. By utilizing multi-camera driving videos, DriveWorld learns spatio-temporal representations, resulting in precise 3D occupancy predictions and addressing both aleatoric and epistemic uncertainties. 

During training, the system learns the state distribution derived from human driving behavior; however, in real-world deployment, it often encounters previously unseen states, leading to degraded model performance or even failure. To address this issue, \cite{mitigating} proposes a latent space generative world model that samples novel states directly from the latent space, enabling the driving policy to learn to recover from off-distribution errors. This approach effectively broadens the scope of the model’s learning, significantly reducing the reliance on large volumes of recovery trajectory data.

\subsubsection{Reinforcement Learning Structure}
Another type of world model is based on the reinforcement learning framework, which utilizes deep neural networks to replace the complex learning processes within its framework. This approach focuses more on theoretical research and how to represent the probability distribution of the real world. Such work typically involves abstract frameworks and tends to adopt an end-to-end application mode. Therefore, this paper classifies it as another structure as Reinforcement Learning Based.

\cite{think2drive} introduces Think2Drive, the first model-based reinforcement learning approach for handling complex, quasi-realistic scenarios in autonomous driving using the CARLA simulator. The method leverages a world model to simulate environmental transitions and train a planner model efficiently. Key contributions include innovative training techniques such as scenario generation and termination-priority replay strategy, a new balanced evaluation metric called Weighted Driving Score, and successful demonstration of expert-level performance in CARLA v2 within three days of training on a single GPU. \cite{sem2} presents the SEMantic Masked recurrent world model (SEM2), which aims to improve the sample efficiency and robustness of end-to-end autonomous driving systems. SEM2 achieves this by incorporating a semantic filter that extracts key driving-relevant features and employs a multi-source data sampler to balance the training data distribution. The experimental results demonstrate that SEM2 outperforms state-of-the-art methods in terms of sample efficiency and robustness against input permutations .

Simultaneously, models like MUVO \cite{muvo} and MILE \cite{mile} showcase the integration of multimodal data and geometric representations to predict future states more accurately. MUVO's use of raw camera and LiDAR data to learn geometric representations underscores the potential for high-resolution environment modeling, crucial for navigating dynamic driving conditions. The key innovation lies in its 3D geometric voxel representations, which allow for precise and actionable environment modeling that supports complex planning and decision-making tasks in autonomous driving. MILE leverages model-based imitation learning to utilize high-resolution videos for offline learning, improving driving outcomes in varied urban scenarios without real-time interaction. This method utilizes 3D lifted features pooled into a bird's-eye view to create a compact, predictive model of urban driving dynamics. Significantly improving upon previous models, MILE demonstrates a 31\% increase in driving score in the CARLA simulator under new town and weather conditions, showcasing its robustness and potential for real-world application.

\section{Datasets and Metrics}

Given the current lack of standardized benchmarks for world models in autonomous driving, this paper emphasizes the importance of datasets and evaluation metrics tailored specifically to this domain. Unlike the earlier discussion on video generation, which will not be revisited here, our focus is directed towards the data resources and metrics that are essential for the progression of world models within autonomous driving, as presented in Table \ref{tab:data}.

\begin{table*}[h]
\centering  
\setlength{\tabcolsep}{4pt} 
\caption{Summaries of the Public Datasets}
\label{tab:data}
\begin{tabularx}{0.95\textwidth}{>{\centering\arraybackslash}Xcc>{\centering\arraybackslash}X>{\centering\arraybackslash}Xc} 

\textbf{Dataset Name} & \textbf{2D} & \textbf{3D} & \textbf{Data Format} & \textbf{Annotation Type} & \textbf{Image Resolution} \\ \hline
nuScenes\cite{nuscenes}  & \Checkmark & \Checkmark & Piont Cloud, Image & segmentation, bbox, tracking & 1600x900             \\ 
KITTI\cite{KITTI}     & \Checkmark & \Checkmark & Piont Cloud, Image & segmentation, bbox           & 1242x375             \\ 
KITTI-360\cite{360kitti} & \Checkmark & \Checkmark & Piont Cloud, Image & segmentation, bbox           & 1242x375             \\ 
Waymo\cite{waymo}     & \Checkmark & \Checkmark & Piont Cloud, Image & segmentation, bbox, keypoint & 1920x1080            \\  
OpenDV-2K\cite{opendv2k} & \Checkmark & \XSolidBrush  & Image, Video       & None                         & 1920x1080            \\ 
Arg2\cite{Arg2, TrustButVerify}      & \Checkmark & \XSolidBrush  & Piont Cloud, Image & segmentation, bbox, keypoint & Multiple Resolutions \\ 
CARLA\cite{carla}     & \Checkmark & \Checkmark & Piont Cloud, Image & segmentation, bbox, tracking & Multiple Resolutions \\ 
OpenScene\cite{yang2023vidar, openscene2023, sima2023_occnet} & \Checkmark & \Checkmark & Piont Cloud, Image & segmentation, bbox, keypoint & Multiple Resolutions \\ 
DrivingDojo\cite{carla}     & \Checkmark & \XSolidBrush & Video & Actions, Text & 1920x1080\\ 
\hline
\end{tabularx}
\end{table*}

\subsection{Datasets}
The development of robust world models in autonomous driving is critically dependent on high-quality datasets that offer diverse and comprehensive scenarios for model training and evaluation. This paper reviews and summarizes the most prominent datasets used in autonomous driving research, highlighting their unique features in terms of scale, annotation types, and application scenarios, thereby forming a complementary set of research resources.

The nuScenes dataset \cite{nuscenes} has garnered significant attention due to its multimodal sensor setup and comprehensive annotations, making it particularly well-suited for core tasks such as object detection, tracking, and path planning. In contrast, the KITTI \cite{KITTI} dataset, an early standard in autonomous driving research, is widely adopted for 2D and 3D scene understanding tasks owing to its simple yet effective structure. Responding to the growing demand for more diverse scenarios, KITTI-360 \cite{360kitti} extends the original dataset with panoramic views and broader scene coverage, enhancing its applicability to tasks such as 3D reconstruction and SLAM.

Waymo \cite{waymo} exemplifies the trend toward large-scale datasets in autonomous driving. With its high-resolution, multi-sensor data and extensive annotations, it serves as a crucial benchmark for perception and prediction tasks in complex urban environments. Meanwhile, the upgraded version of the dataset Argoverse \cite{Argoverse}, Arg2 \cite{Arg2, TrustButVerify}, supports more complex dynamic object tracking and 3D scene reconstruction through more comprehensive annotations and scene expansion.

Datasets such as nuScenes and Waymo are primarily geared towards perception tasks, lacking diverse driving scenarios and comprehensive multi-agent interaction data. In contrast, DrivingDojo \cite{wang2024drivingdojo} is a large-scale driving video dataset specifically designed to advance research in autonomous driving world models. By incorporating a wide variety of driving maneuvers, multi-agent interactions, and open-world knowledge, DrivingDojo lays a solid foundation for the future development of autonomous driving world models.

In contrast to these real-world datasets, CARLA \cite{carla} offers a simulation platform that enables the generation of synthetic data. Its flexibility allows researchers to simulate complex scenarios in a safe and controlled environment, thereby testing the robustness of autonomous driving algorithms.

OpenScene \cite{yang2023vidar, openscene2023, sima2023_occnet} introduces an open-vocabulary framework for 3D scene understanding, breaking the limitations of traditional annotations and enabling zero-shot 3D semantic segmentation and scene understanding. This capability enhances the adaptability of autonomous driving systems to unfamiliar environments.

In conclusion, these datasets collectively provide comprehensive support for autonomous driving research, from fundamental perception tasks to advanced prediction, driving significant technological progress in the field.

\subsection{Metrics}

In the field of autonomous driving, world models are assessed using a diverse set of metrics tailored to specific tasks, including simulation, end-to-end driving, video generation, and 3D scene reconstruction. Due to the heterogeneity of these tasks, no single metric can comprehensively evaluate model performance across all studies. As a result, research in this domain employs a variety of specialized metrics that best capture the accuracy and effectiveness of models within their respective contexts.

For simulation tasks in autonomous driving, prediction accuracy is often evaluated using the Chamfer distance, which quantifies the similarity between generated 3D point clouds and ground truth data, thereby assessing how well a model captures spatial details in scene reconstruction. Consistency, particularly multi-view consistency \cite{nagel1983displacement}, evaluates the spatial and temporal coherence of generated scenes from different perspectives, ensuring logical consistency in the model’s outputs across various viewpoints. Controllability is another key metric \cite{salimans2016improved}, assessed by determining whether the model can generate scenes accurately based on specific input conditions, such as textual descriptions or action commands, thus demonstrating the model’s flexibility and adaptability.

In end-to-end driving tasks, metrics such as the driving score and trajectory deviation are pivotal. The driving score includes measures like the completion rate, which indicates the proportion of driving tasks successfully completed without collisions or deviations from the intended path. Collision rate is critical for assessing model safety, with a lower collision rate reflecting better obstacle avoidance and safer driving decisions. Trajectory deviation measures the difference between predicted and actual driving paths, offering insights into the model’s precision in path planning and execution.

For video and image generation tasks within world models, Fréchet Inception Distance (FID) \cite{heusel2017gans} is widely used. FID measures the distance between the feature distributions of generated and real images or video frames, providing an indication of the visual quality and realism of the generated content. Lower FID scores suggest that the generated outputs are closer to the real data in terms of visual appearance. Structural Similarity Index (SSIM) \cite{wang2004image} is another important metric that evaluates the perceived quality of generated images or frames by comparing them with ground truth data in terms of luminance, contrast, and structural information.

In the context of 3D scene generation, metrics like occupancy grid accuracy and point cloud similarity are essential for assessing the fidelity of generated environments. Occupancy grid accuracy \cite{elfes1989using} compares the predicted occupancy of grid cells in a 3D space with actual occupancy, highlighting the model’s spatial understanding capabilities. Point cloud similarity \cite{besl1992method} measures the closeness of generated point clouds to those derived from real-world LiDAR scans, indicating how accurately the model can reconstruct complex 3D environments.

In summary, these metrics provide a comprehensive framework for evaluating world models across various autonomous driving tasks. They offer detailed insights into the strengths and limitations of models, guiding further refinements and ensuring that the models meet the stringent requirements of real-world applications.

\section{Prospects and Challenges}

World models based methods represent a cutting-edge approach in autonomous driving, enabling high-precision simulation and prediction of future driving scenarios. This technology enhances the situational awareness and decision-making capabilities of autonomous vehicles (AVs) by creating detailed visualizations of potential future events. As this technology continues to evolve, it presents numerous opportunities and challenges.

\subsection{Future Prospects}

\subsubsection{Enhanced Integration of Multimodal Perception and Control}

Future world models aim to integrate various perception and control signals, moving beyond traditional modular designs. Utilizing MLLMs \cite{ofa,llava} and Visual Diffusion Models (VDM) \cite{LDM}, these systems will unify visual and action signals processing. Such models will predict control signals for the current frame from vision-action pairs and forecast future frames based on historical data, thus achieving "Infinite Driving" capabilities. This integration will allow AVs to seamlessly interpret complex traffic environments, react promptly to unexpected obstacles, and optimize their navigation strategies. Moreover, the unification of multimodal data will enable more robust and resilient AV systems capable of maintaining high performance even in challenging conditions such as adverse weather or heavy traffic.

\subsubsection{Diversified Driving Scene Generation}

Advanced world models like DriveDreamer-2 and GAIA-1 utilize generative models to create diverse and realistic driving videos. These videos can be used to train various driving perception methods, improving their effectiveness in real-world applications. These models not only generate high-quality driving videos but also allow customized scene generation through user-friendly text prompts, increasing data diversity and generation quality. This capability is crucial for developing AVs that can handle a wide range of driving conditions, from congested urban streets to remote rural roads. Moreover, by generating diverse scenarios, these models help identify and address potential weaknesses in AV systems, ensuring they are well-prepared for real-world deployment.

\subsubsection{Extended Unsupervised Learning Capabilities}

The next generation of world models will further extend unsupervised learning capabilities like \cite{drivedreamer1, copilot4d}, handling complex observation spaces through discrete diffusion and tokenization techniques, enabling efficient learning in the absence of labels. This will significantly enhance the adaptive and generalization abilities of autonomous driving systems in dynamic environments. Unsupervised learning will allow AVs to continuously improve by learning from the vast amounts of data they encounter during operation, without the need for manual annotation. This approach will reduce development costs and accelerate the deployment of AV technologies. Furthermore, enhanced unsupervised learning will enable AVs to better understand and adapt to new and unforeseen driving situations, improving their overall performance and safety.

\subsection{Challenges}

\subsubsection{Data Scarcity and Annotation Complexity}

Despite significant progress in simulated environments, practical applications still face challenges of data scarcity and complex annotations. Training models that can generalize to real-world complex scenarios requires large amounts of high-quality training data, the collection and annotation of which are often time-consuming and expensive. Furthermore, the variability in driving conditions across different regions necessitates a diverse dataset to ensure comprehensive model training. Addressing these issues involves not only collecting more data but also developing innovative methods for efficient data annotation, such as leveraging semi-supervised learning techniques or using synthetic data to augment real-world datasets \cite{gaia1}.

\subsubsection{Computational Resources and Efficiency}

Training high-precision world models requires substantial computational resources and time. The training of diffusion models \cite{DDIM,DDPM,guided_dm} and LLMs \cite{bert, llama} , in particular, demands high-performance hardware and long training periods, posing a significant barrier for research teams with limited resources. Furthermore, deploying these models in real-time AV systems necessitates optimizing algorithms to run efficiently on limited computational power available on-board. Innovations in hardware acceleration, such as the use of specialized chips for AI processing, and advancements in algorithmic efficiency will be critical to overcoming these challenges and enabling the practical application of advanced world models in autonomous driving.

\subsubsection{Privacy Concerns}

AVs rely heavily on data collection and processing to navigate and interact with their environment. This includes continuous monitoring of their surroundings as well as the collection of passenger data for optimizing travel routes and schedules. Such extensive data collection raises significant privacy concerns, particularly regarding who owns this data, how it is used, and how the privacy of individuals is protected against misuse. Addressing these concerns requires robust data governance frameworks, including clear policies on data ownership, transparent data usage practices, and stringent measures to ensure data security and protect individual privacy. Ensuring public trust in AV technologies will be crucial for their widespread adoption.

\subsubsection{Structural Innovation}

There is an abundance of work based on world models, their overall structures remain largely similar. However, this does not imply that these model structures are already perfected. Fitting neural networks to the real world is a highly challenging task, and thus, the exploration of procedural structures and model selection remains a significant undertaking. Furthermore, current tasks focusing on generating road scenes and video generation still yield suboptimal results, indicating considerable room for improvement.

By continuously overcoming these challenges, future world models will better support the development of autonomous driving technologies, laying the foundation for safer and more efficient autonomous driving systems. This progress will not only revolutionize the way we perceive and interact with transportation but also significantly enhance the overall safety and efficiency of our roadways.

\subsection{Conclusion}

In summary, this review has explored the structural and conceptual similarities between video generation models and world models, underscoring the emerging and still undefined concept of world models. World models have made significant contributions to the field of autonomous driving, particularly in efficiently modeling the real world and accurately predicting driving decisions. Additionally, we have discussed the future prospects and significant challenges in this domain, emphasizing that the structural design and optimization of world models remain ongoing tasks. World models hold immense potential in the autonomous driving sector, offering advanced capabilities for real-world simulation and predictive analysis. Their ability to integrate seamlessly with video generation models enhances the situational awareness and decision-making accuracy of autonomous vehicles. Despite these advancements, several challenges persist, including the need for more sophisticated data integration, efficient model training, and the development of robust frameworks capable of handling complex and diverse driving scenarios. Addressing these challenges will be crucial for the continued evolution and application of world models in autonomous driving, paving the way for safer and more reliable autonomous vehicle systems.

\section*{Acknowledgement}
This work was supported by DiDi GAIA Research Cooperation Initiative (Grant No. CCF-DiDi GAIA 202304).

\ifCLASSOPTIONcaptionsoff
  \newpage
\fi



%
\bibliographystyle{IEEEtran}
\bibliography{IEEEabrv, references}

\begin{thebibliography}{100}
\providecommand{\url}[1]{#1}
\csname url@samestyle\endcsname
\providecommand{\newblock}{\relax}
\providecommand{\bibinfo}[2]{#2}
\providecommand{\BIBentrySTDinterwordspacing}{\spaceskip=0pt\relax}
\providecommand{\BIBentryALTinterwordstretchfactor}{4}
\providecommand{\BIBentryALTinterwordspacing}{\spaceskip=\fontdimen2\font plus
\BIBentryALTinterwordstretchfactor\fontdimen3\font minus \fontdimen4\font\relax}
\providecommand{\BIBforeignlanguage}[2]{{%
\expandafter\ifx\csname l@#1\endcsname\relax
\typeout{** WARNING: IEEEtran.bst: No hyphenation pattern has been}%
\typeout{** loaded for the language `#1'. Using the pattern for}%
\typeout{** the default language instead.}%
\else
\language=\csname l@#1\endcsname
\fi
#2}}
\providecommand{\BIBdecl}{\relax}
\BIBdecl

\bibitem{drivedreamer1}
X.~Wang, Z.~Zhu, G.~Huang, X.~Chen, and J.~Lu, ``Drivedreamer: Towards real-world-driven world models for autonomous driving,'' \emph{arXiv preprint arXiv:2309.09777}, 2023.

\bibitem{drivedreamer2}
G.~Zhao, X.~Wang, Z.~Zhu, X.~Chen, G.~Huang, X.~Bao, and X.~Wang, ``Drivedreamer-2: Llm-enhanced world models for diverse driving video generation,'' \emph{arXiv preprint arXiv:2403.06845}, 2024.

\bibitem{drivingdiffusion}
X.~Li, Y.~Zhang, and X.~Ye, ``Drivingdiffusion: Layout-guided multi-view driving scene video generation with latent diffusion model,'' \emph{arXiv preprint arXiv:2310.07771}, 2023.

\bibitem{adriver}
F.~Jia, W.~Mao, Y.~Liu, Y.~Zhao, Y.~Wen, C.~Zhang, X.~Zhang, and T.~Wang, ``Adriver-i: A general world model for autonomous driving,'' \emph{arXiv preprint arXiv:2311.13549}, 2023.

\bibitem{gaia1}
A.~Hu, L.~Russell, H.~Yeo, Z.~Murez, G.~Fedoseev, A.~Kendall, J.~Shotton, and G.~Corrado, ``Gaia-1: A generative world model for autonomous driving,'' \emph{arXiv preprint arXiv:2309.17080}, 2023.

\bibitem{mile}
A.~Hu, G.~Corrado, N.~Griffiths, Z.~Murez, C.~Gurau, H.~Yeo, A.~Kendall, R.~Cipolla, and J.~Shotton, ``Model-based imitation learning for urban driving,'' \emph{Advances in Neural Information Processing Systems}, vol.~35, pp. 20\,703--20\,716, 2022.

\bibitem{muvo}
D.~Bogdoll, Y.~Yang, and J.~M. Z{\"o}llner, ``Muvo: A multimodal generative world model for autonomous driving with geometric representations,'' \emph{arXiv preprint arXiv:2311.11762}, 2023.

\bibitem{copilot4d}
L.~Zhang, Y.~Xiong, Z.~Yang, S.~Casas, R.~Hu, and R.~Urtasun, ``Learning unsupervised world models for autonomous driving via discrete diffusion,'' \emph{arXiv preprint arXiv:2311.01017}, 2023.

\bibitem{gan}
I.~Goodfellow, J.~Pouget-Abadie, M.~Mirza, B.~Xu, D.~Warde-Farley, S.~Ozair, A.~Courville, and Y.~Bengio, ``Generative adversarial networks,'' \emph{Communications of the ACM}, vol.~63, no.~11, pp. 139--144, 2020.

\bibitem{vae}
D.~P. Kingma, ``Auto-encoding variational bayes,'' \emph{arXiv preprint arXiv:1312.6114}, 2013.

\bibitem{pvdm}
S.~Yu, K.~Sohn, S.~Kim, and J.~Shin, ``Video probabilistic diffusion models in projected latent space,'' in \emph{Proceedings of the IEEE/CVF conference on computer vision and pattern recognition}, 2023, pp. 18\,456--18\,466.

\bibitem{LVDM}
Y.~He, T.~Yang, Y.~Zhang, Y.~Shan, and Q.~Chen, ``Latent video diffusion models for high-fidelity long video generation,'' \emph{arXiv preprint arXiv:2211.13221}, 2022.

\bibitem{videocomposer}
X.~Wang, H.~Yuan, S.~Zhang, D.~Chen, J.~Wang, Y.~Zhang, Y.~Shen, D.~Zhao, and J.~Zhou, ``Videocomposer: Compositional video synthesis with motion controllability,'' \emph{Advances in Neural Information Processing Systems}, vol.~36, 2024.

\bibitem{ho2022video}
J.~Ho, T.~Salimans, A.~Gritsenko, W.~Chan, M.~Norouzi, and D.~J. Fleet, ``Video diffusion models,'' \emph{Advances in Neural Information Processing Systems}, vol.~35, pp. 8633--8646, 2022.

\bibitem{ho2022imagen}
J.~Ho, W.~Chan, C.~Saharia, J.~Whang, R.~Gao, A.~Gritsenko, D.~P. Kingma, B.~Poole, M.~Norouzi, D.~J. Fleet \emph{et~al.}, ``Imagen video: High definition video generation with diffusion models,'' \emph{arXiv preprint arXiv:2210.02303}, 2022.

\bibitem{harvey2022flexible}
W.~Harvey, S.~Naderiparizi, V.~Masrani, C.~Weilbach, and F.~Wood, ``Flexible diffusion modeling of long videos,'' \emph{Advances in Neural Information Processing Systems}, vol.~35, pp. 27\,953--27\,965, 2022.

\bibitem{yu2023video}
S.~Yu, K.~Sohn, S.~Kim, and J.~Shin, ``Video probabilistic diffusion models in projected latent space,'' in \emph{Proceedings of the IEEE/CVF conference on computer vision and pattern recognition}, 2023, pp. 18\,456--18\,466.

\bibitem{tulyakov2017mocogan}
S.~Tulyakov, M.-Y. Liu, X.~Yang, and J.~Kautz, ``Mocogan: Decomposing motion and content for video generation,'' in \emph{Proceedings of the IEEE conference on computer vision and pattern recognition}, 2018, pp. 1526--1535.

\bibitem{yu2022DIGAN}
S.~Yu, J.~Tack, S.~Mo, H.~Kim, J.~Kim, J.-W. Ha, and J.~Shin, ``Generating videos with dynamics-aware implicit generative adversarial networks,'' \emph{arXiv preprint arXiv:2202.10571}, 2022.

\bibitem{survey1}
Z.~Zhu, X.~Wang, W.~Zhao, C.~Min, N.~Deng, M.~Dou, Y.~Wang, B.~Shi, K.~Wang, C.~Zhang \emph{et~al.}, ``Is sora a world simulator? a comprehensive survey on general world models and beyond,'' \emph{arXiv preprint arXiv:2405.03520}, 2024.

\bibitem{survey2}
Y.~Guan, H.~Liao, Z.~Li, J.~Hu, R.~Yuan, Y.~Li, G.~Zhang, and C.~Xu, ``World models for autonomous driving: An initial survey,'' \emph{IEEE Transactions on Intelligent Vehicles}, 2024.

\bibitem{survey3}
C.~Cui, Y.~Ma, X.~Cao, W.~Ye, Y.~Zhou, K.~Liang, J.~Chen, J.~Lu, Z.~Yang, K.-D. Liao \emph{et~al.}, ``A survey on multimodal large language models for autonomous driving,'' in \emph{Proceedings of the IEEE/CVF Winter Conference on Applications of Computer Vision}, 2024, pp. 958--979.

\bibitem{ConvLSTM}
X.~Shi, Z.~Chen, H.~Wang, D.-Y. Yeung, W.-K. Wong, and W.-c. Woo, ``Convolutional lstm network: A machine learning approach for precipitation nowcasting,'' \emph{Advances in neural information processing systems}, vol.~28, 2015.

\bibitem{predrnn}
Y.~Wang, H.~Wu, J.~Zhang, Z.~Gao, J.~Wang, S.~Y. Philip, and M.~Long, ``Predrnn: A recurrent neural network for spatiotemporal predictive learning,'' \emph{IEEE Transactions on Pattern Analysis and Machine Intelligence}, vol.~45, no.~2, pp. 2208--2225, 2022.

\bibitem{videogpt}
W.~Yan, Y.~Zhang, P.~Abbeel, and A.~Srinivas, ``Videogpt: Video generation using vq-vae and transformers,'' \emph{arXiv preprint arXiv:2104.10157}, 2021.

\bibitem{cogvideo}
W.~Hong, M.~Ding, W.~Zheng, X.~Liu, and J.~Tang, ``Cogvideo: Large-scale pretraining for text-to-video generation via transformers,'' \emph{arXiv preprint arXiv:2205.15868}, 2022.

\bibitem{pose}
J.~Walker, K.~Marino, A.~Gupta, and M.~Hebert, ``The pose knows: Video forecasting by generating pose futures,'' in \emph{Proceedings of the IEEE international conference on computer vision}, 2017, pp. 3332--3341.

\bibitem{DDPM}
J.~Ho, A.~Jain, and P.~Abbeel, ``Denoising diffusion probabilistic models,'' \emph{Advances in neural information processing systems}, vol.~33, pp. 6840--6851, 2020.

\bibitem{DDIM}
J.~Song, C.~Meng, and S.~Ermon, ``Denoising diffusion implicit models,'' \emph{arXiv preprint arXiv:2010.02502}, 2020.

\bibitem{saharia2022palette}
C.~Saharia, W.~Chan, H.~Chang, C.~Lee, J.~Ho, T.~Salimans, D.~Fleet, and M.~Norouzi, ``Palette: Image-to-image diffusion models,'' in \emph{ACM SIGGRAPH 2022 conference proceedings}, 2022, pp. 1--10.

\bibitem{xia2023diffir}
B.~Xia, Y.~Zhang, S.~Wang, Y.~Wang, X.~Wu, Y.~Tian, W.~Yang, and L.~Van~Gool, ``Diffir: Efficient diffusion model for image restoration,'' in \emph{Proceedings of the IEEE/CVF International Conference on Computer Vision}, 2023, pp. 13\,095--13\,105.

\bibitem{zhang2023adding}
L.~Zhang, A.~Rao, and M.~Agrawala, ``Adding conditional control to text-to-image diffusion models,'' in \emph{Proceedings of the IEEE/CVF International Conference on Computer Vision}, 2023, pp. 3836--3847.

\bibitem{aboulaich2008new}
R.~Aboulaich, D.~Meskine, and A.~Souissi, ``New diffusion models in image processing,'' \emph{Computers \& Mathematics with Applications}, vol.~56, no.~4, pp. 874--882, 2008.

\bibitem{gu2022vector}
S.~Gu, D.~Chen, J.~Bao, F.~Wen, B.~Zhang, D.~Chen, L.~Yuan, and B.~Guo, ``Vector quantized diffusion model for text-to-image synthesis,'' in \emph{Proceedings of the IEEE/CVF conference on computer vision and pattern recognition}, 2022, pp. 10\,696--10\,706.

\bibitem{LDM}
R.~Rombach, A.~Blattmann, D.~Lorenz, P.~Esser, and B.~Ommer, ``High-resolution image synthesis with latent diffusion models,'' in \emph{Proceedings of the IEEE/CVF conference on computer vision and pattern recognition}, 2022, pp. 10\,684--10\,695.

\bibitem{vqvae}
A.~Van Den~Oord, O.~Vinyals \emph{et~al.}, ``Neural discrete representation learning,'' \emph{Advances in neural information processing systems}, vol.~30, 2017.

\bibitem{DiT}
W.~Peebles and S.~Xie, ``Scalable diffusion models with transformers,'' in \emph{Proceedings of the IEEE/CVF International Conference on Computer Vision}, 2023, pp. 4195--4205.

\bibitem{magvit}
L.~Yu, Y.~Cheng, K.~Sohn, J.~Lezama, H.~Zhang, H.~Chang, A.~G. Hauptmann, M.-H. Yang, Y.~Hao, I.~Essa \emph{et~al.}, ``Magvit: Masked generative video transformer,'' in \emph{Proceedings of the IEEE/CVF Conference on Computer Vision and Pattern Recognition}, 2023, pp. 10\,459--10\,469.

\bibitem{maskgit}
F.~Ebert, C.~Finn, A.~X. Lee, and S.~Levine, ``Self-supervised visual planning with temporal skip connections.'' \emph{CoRL}, vol.~12, no.~16, p.~23, 2017.

\bibitem{worldmodel}
D.~Ha and J.~Schmidhuber, ``World models,'' \emph{arXiv preprint arXiv:1803.10122}, 2018.

\bibitem{MDN}
C.~M. Bishop, ``Mixture density networks,'' 1994.

\bibitem{RNN}
S.~Grossberg, ``Recurrent neural networks,'' \emph{Scholarpedia}, vol.~8, no.~2, p. 1888, 2013.

\bibitem{PlaNet}
D.~Hafner, T.~Lillicrap, I.~Fischer, R.~Villegas, D.~Ha, H.~Lee, and J.~Davidson, ``Learning latent dynamics for planning from pixels,'' in \emph{International conference on machine learning}.\hskip 1em plus 0.5em minus 0.4em\relax PMLR, 2019, pp. 2555--2565.

\bibitem{Dreamer1}
D.~Hafner, T.~Lillicrap, J.~Ba, and M.~Norouzi, ``Dream to control: Learning behaviors by latent imagination,'' \emph{arXiv preprint arXiv:1912.01603}, 2019.

\bibitem{Dreamer2}
D.~Hafner, T.~Lillicrap, M.~Norouzi, and J.~Ba, ``Mastering atari with discrete world models,'' \emph{arXiv preprint arXiv:2010.02193}, 2020.

\bibitem{Dreamer3}
D.~Hafner, J.~Pasukonis, J.~Ba, and T.~Lillicrap, ``Mastering diverse domains through world models,'' \emph{arXiv preprint arXiv:2301.04104}, 2023.

\bibitem{lecun2022path}
Y.~LeCun, ``A path towards autonomous machine intelligence version 0.9. 2, 2022-06-27,'' \emph{Open Review}, vol.~62, no.~1, 2022.

\bibitem{ijepa}
M.~Assran, Q.~Duval, I.~Misra, P.~Bojanowski, P.~Vincent, M.~Rabbat, Y.~LeCun, and N.~Ballas, ``Self-supervised learning from images with a joint-embedding predictive architecture,'' in \emph{Proceedings of the IEEE/CVF Conference on Computer Vision and Pattern Recognition}, 2023, pp. 15\,619--15\,629.

\bibitem{vjepa}
A.~Bardes, Q.~Garrido, J.~Ponce, X.~Chen, M.~Rabbat, Y.~LeCun, M.~Assran, and N.~Ballas, ``Revisiting feature prediction for learning visual representations from video,'' \emph{arXiv preprint arXiv:2404.08471}, 2024.

\bibitem{IWM}
Q.~Garrido, M.~Assran, N.~Ballas, A.~Bardes, L.~Najman, and Y.~LeCun, ``Learning and leveraging world models in visual representation learning,'' \emph{arXiv preprint arXiv:2403.00504}, 2024.

\bibitem{sora}
\BIBentryALTinterwordspacing
B.~Tim, P.~Bill, and H.~Connor, et~al., ``Video generation models as world simulators,'' 2024. [Online]. Available: \url{https://openai.com/research/video-generation-models-as-world-simulators}
\BIBentrySTDinterwordspacing

\bibitem{genie}
J.~Bruce, M.~D. Dennis, A.~Edwards, J.~Parker-Holder, Y.~Shi, E.~Hughes, M.~Lai, A.~Mavalankar, R.~Steigerwald, C.~Apps \emph{et~al.}, ``Genie: Generative interactive environments,'' in \emph{Forty-first International Conference on Machine Learning}, 2024.

\bibitem{scenediffusion}
E.~Pronovost, K.~Wang, and N.~Roy, ``Generating driving scenes with diffusion,'' \emph{arXiv preprint arXiv:2305.18452}, 2023.

\bibitem{panacea}
Y.~Wen, Y.~Zhao, and Y.~Liu, et~al., ``Panacea: Panoramic and controllable video generation for autonomous driving,'' in \emph{Proceedings of the IEEE/CVF Conference on Computer Vision and Pattern Recognition}, 2024, pp. 6902--6912.

\bibitem{contorlNet}
L.~Zhang, A.~Rao, and M.~Agrawala, ``Adding conditional control to text-to-image diffusion models,'' in \emph{Proceedings of the IEEE/CVF International Conference on Computer Vision}, 2023, pp. 3836--3847.

\bibitem{GenAD1}
J.~Yang, S.~Gao, Y.~Qiu, L.~Chen, T.~Li, B.~Dai, K.~Chitta, P.~Wu, J.~Zeng, P.~Luo \emph{et~al.}, ``Generalized predictive model for autonomous driving,'' in \emph{Proceedings of the IEEE/CVF Conference on Computer Vision and Pattern Recognition}, 2024, pp. 14\,662--14\,672.

\bibitem{Delphi}
E.~Ma, L.~Zhou, T.~Tang, Z.~Zhang, D.~Han, J.~Jiang, K.~Zhan, P.~Jia, X.~Lang, H.~Sun \emph{et~al.}, ``Unleashing generalization of end-to-end autonomous driving with controllable long video generation,'' \emph{arXiv preprint arXiv:2406.01349}, 2024.

\bibitem{Vista}
S.~Gao, J.~Yang, L.~Chen, K.~Chitta, Y.~Qiu, A.~Geiger, J.~Zhang, and H.~Li, ``Vista: A generalizable driving world model with high fidelity and versatile controllability,'' \emph{arXiv preprint arXiv:2405.17398}, 2024.

\bibitem{OccSora}
L.~Wang, W.~Zheng, Y.~Ren, H.~Jiang, Z.~Cui, H.~Yu, and J.~Lu, ``Occsora: 4d occupancy generation models as world simulators for autonomous driving,'' \emph{arXiv preprint arXiv:2405.20337}, 2024.

\bibitem{BEVWorld}
Y.~Zhang, S.~Gong, K.~Xiong, X.~Ye, X.~Tan, F.~Wang, J.~Huang, H.~Wu, and H.~Wang, ``Bevworld: A multimodal world model for autonomous driving via unified bev latent space,'' \emph{arXiv preprint arXiv:2407.05679}, 2024.

\bibitem{vidar}
Z.~Yang, L.~Chen, Y.~Sun, and H.~Li, ``Visual point cloud forecasting enables scalable autonomous driving,'' in \emph{Proceedings of the IEEE/CVF Conference on Computer Vision and Pattern Recognition}, 2024, pp. 14\,673--14\,684.

\bibitem{wu2023next}
S.~Wu, H.~Fei, L.~Qu, W.~Ji, and T.-S. Chua, ``Next-gpt: Any-to-any multimodal llm,'' \emph{arXiv preprint arXiv:2309.05519}, 2023.

\bibitem{ma2023llm}
X.~Ma, G.~Fang, and X.~Wang, ``Llm-pruner: On the structural pruning of large language models,'' \emph{Advances in neural information processing systems}, vol.~36, pp. 21\,702--21\,720, 2023.

\bibitem{ni2023lever}
A.~Ni, S.~Iyer, D.~Radev, V.~Stoyanov, W.-t. Yih, S.~Wang, and X.~V. Lin, ``Lever: Learning to verify language-to-code generation with execution,'' in \emph{International Conference on Machine Learning}.\hskip 1em plus 0.5em minus 0.4em\relax PMLR, 2023, pp. 26\,106--26\,128.

\bibitem{vaithilingam2022expectation}
P.~Vaithilingam, T.~Zhang, and E.~L. Glassman, ``Expectation vs. experience: Evaluating the usability of code generation tools powered by large language models,'' in \emph{Chi conference on human factors in computing systems extended abstracts}, 2022, pp. 1--7.

\bibitem{ding2023static}
H.~Ding, V.~Kumar, Y.~Tian, Z.~Wang, R.~Kwiatkowski, X.~Li, M.~K. Ramanathan, B.~Ray, P.~Bhatia, S.~Sengupta \emph{et~al.}, ``A static evaluation of code completion by large language models,'' \emph{arXiv preprint arXiv:2306.03203}, 2023.

\bibitem{gu2023llm}
Q.~Gu, ``Llm-based code generation method for golang compiler testing,'' in \emph{Proceedings of the 31st ACM Joint European Software Engineering Conference and Symposium on the Foundations of Software Engineering}, 2023, pp. 2201--2203.

\bibitem{llama}
H.~Touvron, T.~Lavril, G.~Izacard, X.~Martinet, M.-A. Lachaux, T.~Lacroix, B.~Rozi{\`e}re, N.~Goyal, E.~Hambro, F.~Azhar \emph{et~al.}, ``Llama: Open and efficient foundation language models,'' \emph{arXiv preprint arXiv:2302.13971}, 2023.

\bibitem{llama2}
M.~GenAI, ``Llama 2: Open foundation and fine-tuned chat models,'' \emph{arXiv preprint arXiv:2307.09288}, 2023.

\bibitem{vicuna}
\BIBentryALTinterwordspacing
W.-L. Chiang, Z.~Li, and Z.~Lin, et~al., ``Vicuna: An open-source chatbot impressing gpt-4 with 90\%* chatgpt quality,'' March 2023. [Online]. Available: \url{https://lmsys.org/blog/2023-03-30-vicuna/}
\BIBentrySTDinterwordspacing

\bibitem{vicuna_}
B.~Peng, C.~Li, P.~He, M.~Galley, and J.~Gao, ``Instruction tuning with gpt-4,'' \emph{arXiv preprint arXiv:2304.03277}, 2023.

\bibitem{gpt}
A.~Radford, K.~Narasimhan, T.~Salimans, I.~Sutskever \emph{et~al.}, ``Improving language understanding by generative pre-training,'' 2018.

\bibitem{gpt2}
A.~Radford, J.~Wu, R.~Child, D.~Luan, D.~Amodei, I.~Sutskever \emph{et~al.}, ``Language models are unsupervised multitask learners,'' \emph{OpenAI blog}, vol.~1, no.~8, p.~9, 2019.

\bibitem{gpt3}
------, ``Language models are unsupervised multitask learners,'' \emph{OpenAI blog}, vol.~1, no.~8, p.~9, 2019.

\bibitem{floridi2020gpt}
L.~Floridi and M.~Chiriatti, ``Gpt-3: Its nature, scope, limits, and consequences,'' \emph{Minds and Machines}, vol.~30, pp. 681--694, 2020.

\bibitem{llava}
H.~Liu, C.~Li, Q.~Wu, and Y.~J. Lee, ``Visual instruction tuning,'' \emph{Advances in neural information processing systems}, vol.~36, 2024.

\bibitem{blip2}
J.~Li, D.~Li, S.~Savarese, and S.~Hoi, ``Blip-2: Bootstrapping language-image pre-training with frozen image encoders and large language models,'' in \emph{International conference on machine learning}.\hskip 1em plus 0.5em minus 0.4em\relax PMLR, 2023, pp. 19\,730--19\,742.

\bibitem{gpt4}
J.~Achiam, S.~Adler, and S.~Agarwal, et~al., ``Gpt-4 technical report,'' \emph{arXiv preprint arXiv:2303.08774}, 2023.

\bibitem{drivegpt4}
Z.~Xu, Y.~Zhang, E.~Xie, Z.~Zhao, Y.~Guo, K.-Y.~K. Wong, Z.~Li, and H.~Zhao, ``Drivegpt4: Interpretable end-to-end autonomous driving via large language model,'' \emph{IEEE Robotics and Automation Letters}, 2024.

\bibitem{wei2024occllama}
J.~Wei, S.~Yuan, P.~Li, Q.~Hu, Z.~Gan, and W.~Ding, ``Occllama: An occupancy-language-action generative world model for autonomous driving,'' \emph{arXiv preprint arXiv:2409.03272}, 2024.

\bibitem{occworld}
W.~Zheng, W.~Chen, Y.~Huang, B.~Zhang, Y.~Duan, and J.~Lu, ``Occworld: Learning a 3d occupancy world model for autonomous driving,'' \emph{arXiv preprint arXiv:2311.16038}, 2023.

\bibitem{driveoccworld}
Y.~Yang, J.~Mei, Y.~Ma, S.~Du, W.~Chen, Y.~Qian, Y.~Feng, and Y.~Liu, ``Driving in the occupancy world: Vision-centric 4d occupancy forecasting and planning via world models for autonomous driving,'' \emph{arXiv preprint arXiv:2408.14197}, 2024.

\bibitem{uniworld}
C.~Min, D.~Zhao, L.~Xiao, Y.~Nie, and B.~Dai, ``Uniworld: Autonomous driving pre-training via world models,'' \emph{arXiv preprint arXiv:2308.07234}, 2023.

\bibitem{nerf}
B.~Mildenhall, P.~P. Srinivasan, M.~Tancik, J.~T. Barron, R.~Ramamoorthi, and R.~Ng, ``Nerf: Representing scenes as neural radiance fields for view synthesis,'' \emph{Communications of the ACM}, vol.~65, no.~1, pp. 99--106, 2021.

\bibitem{3d}
B.~Kerbl, G.~Kopanas, T.~Leimk{\"u}hler, and G.~Drettakis, ``3d gaussian splatting for real-time radiance field rendering.'' \emph{ACM Trans. Graph.}, vol.~42, no.~4, pp. 139--1, 2023.

\bibitem{drivedreamer4d}
G.~Zhao, C.~Ni, X.~Wang, Z.~Zhu, G.~Huang, X.~Chen, B.~Wang, Y.~Zhang, W.~Mei, and X.~Wang, ``Drivedreamer4d: World models are effective data machines for 4d driving scene representation,'' \emph{arXiv preprint arXiv:2410.13571}, 2024.

\bibitem{LAW}
Y.~Li, L.~Fan, J.~He, Y.~Wang, Y.~Chen, Z.~Zhang, and T.~Tan, ``Enhancing end-to-end autonomous driving with latent world model,'' \emph{arXiv preprint arXiv:2406.08481}, 2024.

\bibitem{mitigating}
A.~Popov, A.~Degirmenci, D.~Wehr, S.~Hegde, R.~Oldja, A.~Kamenev, B.~Douillard, D.~Nist{\'e}r, U.~Muller, R.~Bhargava \emph{et~al.}, ``Mitigating covariate shift in imitation learning for autonomous vehicles using latent space generative world models,'' \emph{arXiv preprint arXiv:2409.16663}, 2024.

\bibitem{UnO}
B.~Agro, Q.~Sykora, S.~Casas, T.~Gilles, and R.~Urtasun, ``Uno: Unsupervised occupancy fields for perception and forecasting,'' in \emph{Proceedings of the IEEE/CVF Conference on Computer Vision and Pattern Recognition}, 2024, pp. 14\,487--14\,496.

\bibitem{DriveWorld}
C.~Min, D.~Zhao, L.~Xiao, J.~Zhao, X.~Xu, Z.~Zhu, L.~Jin, J.~Li, Y.~Guo, J.~Xing \emph{et~al.}, ``Driveworld: 4d pre-trained scene understanding via world models for autonomous driving,'' in \emph{Proceedings of the IEEE/CVF Conference on Computer Vision and Pattern Recognition}, 2024, pp. 15\,522--15\,533.

\bibitem{think2drive}
Q.~Li, X.~Jia, S.~Wang, and J.~Yan, ``Think2drive: Efficient reinforcement learning by thinking in latent world model for quasi-realistic autonomous driving (in carla-v2),'' \emph{arXiv preprint arXiv:2402.16720}, 2024.

\bibitem{sem2}
Z.~Gao, Y.~Mu, C.~Chen, J.~Duan, P.~Luo, Y.~Lu, and S.~E. Li, ``Enhance sample efficiency and robustness of end-to-end urban autonomous driving via semantic masked world model,'' \emph{IEEE Transactions on Intelligent Transportation Systems}, 2024.

\bibitem{nuscenes}
H.~Caesar, V.~Bankiti, A.~H. Lang, S.~Vora, V.~E. Liong, Q.~Xu, A.~Krishnan, Y.~Pan, G.~Baldan, and O.~Beijbom, ``nuscenes: A multimodal dataset for autonomous driving,'' in \emph{Proceedings of the IEEE/CVF conference on computer vision and pattern recognition}, 2020, pp. 11\,621--11\,631.

\bibitem{KITTI}
A.~Geiger, P.~Lenz, and R.~Urtasun, ``Are we ready for autonomous driving? the kitti vision benchmark suite,'' in \emph{2012 IEEE conference on computer vision and pattern recognition}.\hskip 1em plus 0.5em minus 0.4em\relax IEEE, 2012, pp. 3354--3361.

\bibitem{360kitti}
Y.~Liao, J.~Xie, and A.~Geiger, ``Kitti-360: A novel dataset and benchmarks for urban scene understanding in 2d and 3d,'' \emph{IEEE Transactions on Pattern Analysis and Machine Intelligence}, vol.~45, no.~3, pp. 3292--3310, 2022.

\bibitem{waymo}
P.~Sun, H.~Kretzschmar, X.~Dotiwalla, A.~Chouard, V.~Patnaik, P.~Tsui, J.~Guo, Y.~Zhou, Y.~Chai, B.~Caine \emph{et~al.}, ``Scalability in perception for autonomous driving: Waymo open dataset,'' in \emph{Proceedings of the IEEE/CVF conference on computer vision and pattern recognition}, 2020, pp. 2446--2454.

\bibitem{opendv2k}
J.~Yang, S.~Gao, Y.~Qiu, L.~Chen, T.~Li, B.~Dai, K.~Chitta, P.~Wu, J.~Zeng, P.~Luo \emph{et~al.}, ``Generalized predictive model for autonomous driving,'' in \emph{Proceedings of the IEEE/CVF Conference on Computer Vision and Pattern Recognition}, 2024, pp. 14\,662--14\,672.

\bibitem{Arg2}
B.~Wilson, W.~Qi, T.~Agarwal, J.~Lambert, J.~Singh, S.~Khandelwal, B.~Pan, R.~Kumar, A.~Hartnett, J.~K. Pontes \emph{et~al.}, ``Argoverse 2: Next generation datasets for self-driving perception and forecasting,'' \emph{arXiv preprint arXiv:2301.00493}, 2023.

\bibitem{TrustButVerify}
J.~Lambert and J.~Hays, ``Trust, but verify: Cross-modality fusion for hd map change detection,'' in \emph{Proceedings of the Neural Information Processing Systems Track on Datasets and Benchmarks (NeurIPS Datasets and Benchmarks 2021)}, 2021.

\bibitem{carla}
A.~Dosovitskiy, G.~Ros, F.~Codevilla, A.~Lopez, and V.~Koltun, ``Carla: An open urban driving simulator,'' in \emph{Conference on robot learning}.\hskip 1em plus 0.5em minus 0.4em\relax PMLR, 2017, pp. 1--16.

\bibitem{yang2023vidar}
Z.~Yang, L.~Chen, Y.~Sun, and H.~Li, ``Visual point cloud forecasting enables scalable autonomous driving,'' \emph{arXiv preprint arXiv:2312.17655}, 2023.

\bibitem{openscene2023}
O.~Contributors, ``Openscene: The largest up-to-date 3d occupancy prediction benchmark in autonomous driving,'' \url{https://github.com/OpenDriveLab/OpenScene}, 2023.

\bibitem{sima2023_occnet}
C.~Sima, W.~Tong, T.~Wang, L.~Chen, S.~Wu, H.~Deng, Y.~Gu, L.~Lu, P.~Luo, D.~Lin, and H.~Li, ``Scene as occupancy,'' 2023.

\bibitem{Argoverse}
M.-F. Chang, J.~W. Lambert, P.~Sangkloy, J.~Singh, S.~Bak, A.~Hartnett, D.~Wang, P.~Carr, S.~Lucey, D.~Ramanan, and J.~Hays, ``Argoverse: 3d tracking and forecasting with rich maps,'' in \emph{Conference on Computer Vision and Pattern Recognition (CVPR)}, 2019.

\bibitem{wang2024drivingdojo}
Y.~Wang, K.~Cheng, J.~He, Q.~Wang, H.~Dai, Y.~Chen, F.~Xia, and Z.~Zhang, ``Drivingdojo dataset: Advancing interactive and knowledge-enriched driving world model,'' \emph{arXiv preprint arXiv:2410.10738}, 2024.

\bibitem{nagel1983displacement}
H.-H. Nagel, ``Displacement vectors derived from second-order intensity variations in image sequences,'' \emph{Computer Vision, Graphics, and Image Processing}, vol.~21, no.~1, pp. 85--117, 1983.

\bibitem{salimans2016improved}
T.~Salimans, I.~Goodfellow, W.~Zaremba, V.~Cheung, A.~Radford, and X.~Chen, ``Improved techniques for training gans,'' \emph{Advances in neural information processing systems}, vol.~29, 2016.

\bibitem{heusel2017gans}
M.~Heusel, H.~Ramsauer, T.~Unterthiner, B.~Nessler, and S.~Hochreiter, ``Gans trained by a two time-scale update rule converge to a local nash equilibrium,'' \emph{Advances in neural information processing systems}, vol.~30, 2017.

\bibitem{wang2004image}
Z.~Wang, A.~C. Bovik, H.~R. Sheikh, and E.~P. Simoncelli, ``Image quality assessment: from error visibility to structural similarity,'' \emph{IEEE transactions on image processing}, vol.~13, no.~4, pp. 600--612, 2004.

\bibitem{elfes1989using}
A.~Elfes, ``Using occupancy grids for mobile robot perception and navigation,'' \emph{Computer}, vol.~22, no.~6, pp. 46--57, 1989.

\bibitem{besl1992method}
P.~J. Besl and N.~D. McKay, ``Method for registration of 3-d shapes,'' in \emph{Sensor fusion IV: control paradigms and data structures}, vol. 1611.\hskip 1em plus 0.5em minus 0.4em\relax Spie, 1992, pp. 586--606.

\bibitem{ofa}
P.~Wang, A.~Yang, R.~Men, J.~Lin, S.~Bai, Z.~Li, J.~Ma, C.~Zhou, J.~Zhou, and H.~Yang, ``Ofa: Unifying architectures, tasks, and modalities through a simple sequence-to-sequence learning framework,'' in \emph{International conference on machine learning}.\hskip 1em plus 0.5em minus 0.4em\relax PMLR, 2022, pp. 23\,318--23\,340.

\bibitem{guided_dm}
P.~Dhariwal and A.~Nichol, ``Diffusion models beat gans on image synthesis,'' \emph{Advances in neural information processing systems}, vol.~34, pp. 8780--8794, 2021.

\bibitem{bert}
J.~Devlin, ``Bert: Pre-training of deep bidirectional transformers for language understanding,'' \emph{arXiv preprint arXiv:1810.04805}, 2018.

\end{thebibliography}
\vspace{-12 mm}

\end{document}